\documentclass[letterpaper, 10 pt, conference]{ieeeconf}  

\usepackage{cleveref}

\IEEEoverridecommandlockouts                              

\usepackage{graphics} 
\usepackage{epsfig} 
\usepackage{mathptmx} 
\usepackage{times} 

\usepackage{amsmath} 
\usepackage{breqn}
\usepackage{amssymb}  
\usepackage[dvipsnames]{xcolor}
\usepackage[noadjust]{cite}
\usepackage[normalem]{ulem}
\usepackage{soul}
\usepackage{booktabs}
\usepackage{nicefrac}
\usepackage{siunitx}
\usepackage{ulem}

\usepackage{algorithm}
\usepackage{algpseudocode}
\algtext*{EndIf}

\newcommand{\eg}{e.g.\ }

\newcommand{\cE}{\mathcal{E}}
\newcommand{\cV}{\mathcal{V}}

\newcommand{\R}{\mathbb{R}}
\newcommand{\Zp}{\mathbb{Z}^+}
\newcommand{\bD}{\mathbf{D}}

\DeclareMathOperator*{\minimize}{minimize\ }

\newtheorem{theorem}{Theorem}[section]
\newtheorem{problem}[theorem]{Problem}

\title{\LARGE \bf
	Resilience by Reconfiguration: Exploiting Heterogeneity in Robot Teams}

\author{Ragesh K. Ramachandran, James A. Preiss and Gaurav S. Sukhatme
	\thanks{This work was supported in part by the Army Research Laboratory as part of the Distributed and Collaborative Intelligent Systems and Technology (DCIST) Collaborative Research Alliance (CRA). The authors are with the Department of Computer Science, University of Southern California, Los Angeles, CA 90089, USA  {\tt\small rageshku|japreiss|gaurav@usc.edu}}
}

\let\oldtitle\title
\renewcommand\title[1]{%
	\begingroup
	\providecommand{\ttlit}{}%
	\renewcommand{\ttlit}[1]{}%
	\providecommand{\titlenote}{}%
	\renewcommand{\titlenote}[1]{}%
	\hypersetup{pdftitle={#1}}%
	\def\thetitle{#1}%
	\pdfbookmark[0]{#1}{title}
	\endgroup
	\oldtitle{#1}%
}

\usepackage{calc}

\usepackage[shortcuts]{extdash}

\usepackage{graphicx}

\usepackage{booktabs}

\usepackage{verbatim}

\usepackage{amsmath}
\usepackage{psfrag}

\makeatletter
\def\citet{\@ifstar{\citetstar}{\citetnostar}}
\def\Citet{\@ifstar{\Citetstar}{\Citetnostar}}
\def\citetnostar{\@ifnextchar[{\squarecitet}{\simplecitet}}
\def\squarecitet[#1]{\@ifnextchar[{\twocitet[#1]}{\onecitet[#1]}}
\def\Citetnostar{\@ifnextchar[{\squareCitet}{\simpleCitet}}
\def\squareCitet[#1]{\@ifnextchar[{\twoCitet[#1]}{\oneCitet[#1]}}
\def\citetstar{\@ifnextchar[{\squarecitetstar}{\simplecitetstar}}
\def\squarecitetstar[#1]{\@ifnextchar[{\twocitetstar[#1]}{\onecitetstar[#1]}}
\def\Citetstar{\@ifnextchar[{\squareCitetstar}{\simpleCitetstar}}
\def\squareCitetstar[#1]{\@ifnextchar[{\twoCitetstar[#1]}{\oneCitetstar[#1]}}
\makeatother
\def\simplecitet#1{\citeauthor{#1}~\citep{#1}}
\def\onecitet[#1]#2{\citeauthor{#2}~\citep[#1]{#2}}
\def\twocitet[#1][#2]#3{\citeauthor{#3}~\citep[#1][#2]{#3}}
\def\simplecitetstar#1{\citeauthor*{#1}~\citep{#1}}
\def\onecitetstar[#1]#2{\citeauthor*{#2}~\citep[#1]{#2}}
\def\twocitetstar[#1][#2]#3{\citeauthor*{#3}~\citep[#1][#2]{#3}}
\def\simpleCitet#1{\Citeauthor{#1}~\citep{#1}}
\def\oneCitet[#1]#2{\Citeauthor{#2}~\citep[#1]{#2}}
\def\twoCitet[#1][#2]#3{\Citeauthor{#3}~\citep[#1][#2]{#3}}
\def\simpleCitetstar#1{\Citeauthor*{#1}~\citep{#1}}
\def\oneCitetstar[#1]#2{\Citeauthor*{#2}~\citep[#1]{#2}}
\def\twoCitetstar[#1][#2]#3{\Citeauthor*{#3}~\citep[#1][#2]{#3}}

\usepackage[cmex10]{mathtools}             %
\usepackage{amsfonts,amssymb}
\usepackage{mathrsfs}                      

\usepackage{varioref}
\labelformat{subfigure}{\thefigure\textup{(#1)}}
\labelformat{equation}{\textup{(#1)}}

\usepackage[%
pagebackref=false,
bookmarks=true,bookmarksopen=true,
bookmarksnumbered=true,
breaklinks=true,
colorlinks=true,
anchorcolor=blue,citecolor=blue,
urlcolor=blue,linkcolor=blue,filecolor=blue,
menucolor=blue,
]{hyperref}
\usepackage{doi}

\usepackage{subcaption}

\makeatletter
\@ifpackageloaded{hyperref}{%
	\@ifpackageloaded{amsmath}{%
		\newcommand{\AMShreffix}[1]{%
			\expandafter\let\csname AMShreffix#1\expandafter\endcsname%
			\csname #1\endcsname%
			\expandafter\renewcommand\csname #1\endcsname{%
				\@hyper@itemfalse\csname AMShreffix#1\endcsname}}
		\AtBeginDocument{%
			\AMShreffix{equation}
			\AMShreffix{align}
			\AMShreffix{alignat}
			\AMShreffix{flalign}
			\AMShreffix{gather}
			\AMShreffix{multline}
	}}{}}{}
\makeatother

\begin{document}
	
	\maketitle
	\thispagestyle{empty}
	\pagestyle{empty}

	\begin{abstract}
		
		We propose a method to maintain high resource availability in a networked heterogeneous multi-robot system subject to resource failures. In our model, resources such as sensing and computation are available on robots. The robots are engaged in a joint task using these pooled resources. When a resource on a particular robot becomes unavailable (e.g., a sensor ceases to function), the system automatically reconfigures so that the robot continues to have access to this resource by communicating with other robots. 
		Specifically, we consider the problem of selecting edges to be modified in the system's communication graph after a resource failure has occurred. We define a metric that allows us to characterize the quality of the resource distribution in the network represented by the communication graph. Upon a resource becoming unavailable due to failure, we reconfigure the network so that the resource distribution is brought as close to the maximal resource distribution as possible without a large change in the 
		number of active inter-robot communication links.
		Our approach uses mixed integer semi-definite programming to achieve this goal.
		We employ a simulated annealing method to compute a spatial formation that satisfies the inter-robot distances imposed by the topology, along with other constraints.
		Our method can compute a communication topology, spatial formation, and formation change motion planning in a few seconds.
		We validate our method in simulation and real-robot experiments with a team of seven quadrotors.
	\end{abstract}
	

	
	\section{Introduction and Related Work}
	\label{sec:intro}
	
	A heterogeneous multi-robot team, where robots have varied sensing, actuation, communication and computational capabilities, is a promising direction to consider when building a resilient team. Such a team can work together by sharing resources between individual robots to perform complex tasks, thereby being resilient to failures of individual robots. For example, when a particular sensor on a robot fails, it may be able to rely on measurements made by a teammate nearby. 
	
	
	If the team is to perform its task by sharing resources between team members, it is desirable to configure the team with a communication topology such that each robot has access to its teammates' resources (e.g., sensing, computation) within some neighborhood. Under distance-limited communication, specifying such a topology restricts the feasible set of the robots' physical locations. 
	
	\begin{figure}
		\centering
		\includegraphics[height=.35\linewidth]{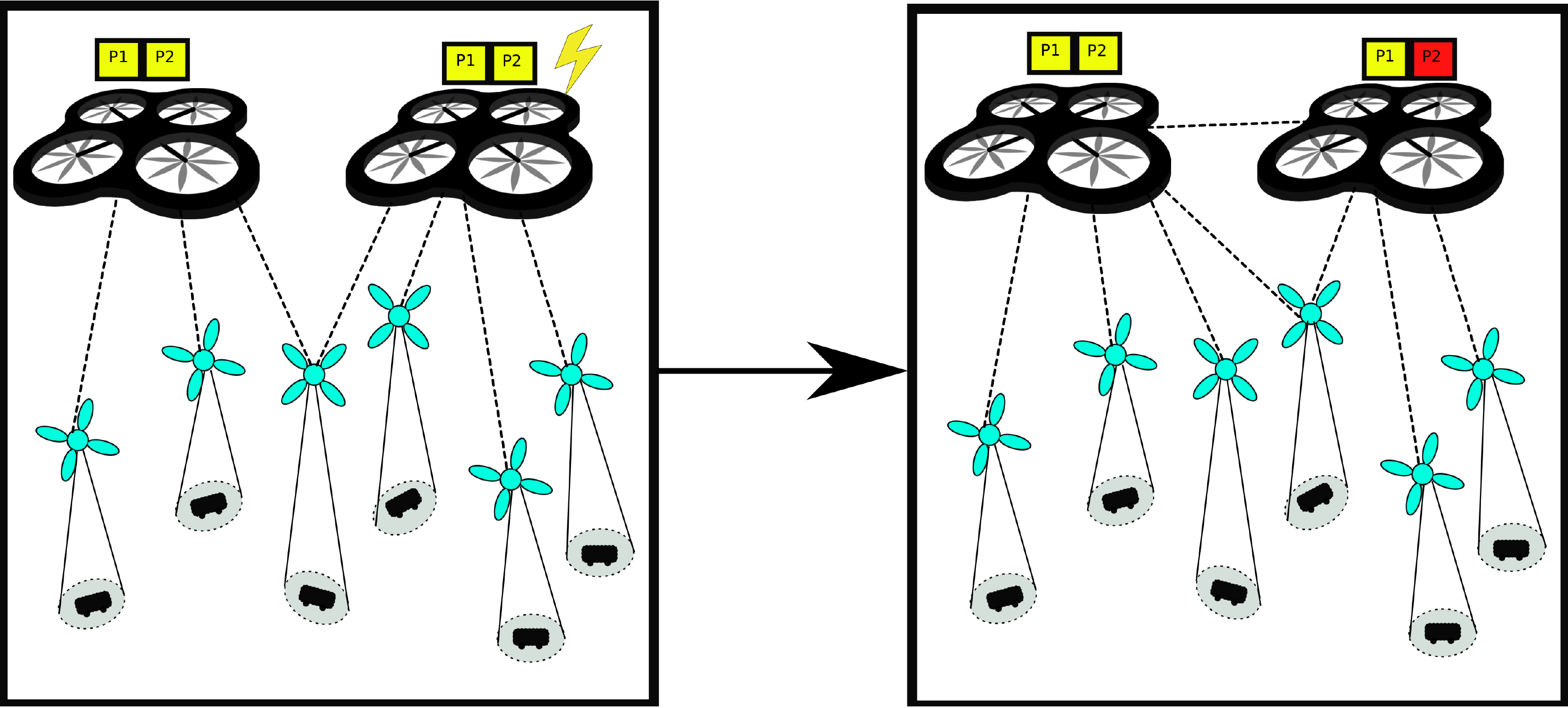}
		\caption{Reconfiguration of communication network topology in a target tracking scenario (quadrotor image: Wikipedia).}
		\vspace{-9mm}
		\label{fig:failure eg}       
	\end{figure}
	
	As motivation, consider a multi-target tracking application by a team with two kinds of quadrotors  as shown in \autoref{fig:failure eg}.  The smaller quadrotors are equipped with sensors (\eg cameras) required to sense and identify targets to be tracked. Each large quadrotor has two kinds of computational resource $P1$ and $P2$ (e.g. GPU and CPU).  Smaller quadrotors do not have the computational resources to carry out the computation required for tracking the target of interest. They rely on the larger quadrotors to process their sensor data. As shown \autoref{fig:failure eg}, if resource $P2$ of a quadrotor fails then the communication network must reconfigure so that $P2$ is available to every member in the team in a few hops. 
	
	Here, we propose a two-stage method to reduce the impact of a resource failure in a heterogeneous robot team engaged in a task. In response to a resource failure, our method first updates the communication network topology such that the new communication topology is close to the original communication topology and the resource distribution within the network is as close as possible to the maximal resource distribution. An maximal resource distribution is one in which  each robot has access to all available resources within a single hop distance. This stage is
	followed by computing a spatial formation of the robots under which the new topology can be achieved by distance-constrained communications. The framework developed here only accounts for the failure of an individual resource at a time (\eg a single sensor) on a robot, not for the case where the robot itself completely fails to function (\eg a quadrotor loses the ability to perform low-level flight control). Thus in our model, when a robot loses its own resources to perform its part in the joint team task, it may still execute its role by relying on the resources of 1-hop neighbors. This leaves open the possibility that in some extreme cases a robot can still help the team perform its overall task \eg even when all its own sensors needed for the team task have failed but it has access to the relevant sensor information from neighbors. In full generality, we are interested in studying the problem where a robot that has lost a resource needed for performing its part in the team task can access the lost resource from other robots in the team with low communication and physical reconfiguration costs. The 1-hop neighbor case is our first attempt to develop an abstraction to concurrently formalize these ideas. In future work, we plan to study the general case where the access "distance" (in hops) to resources is larger than 1.
	
	In stage one, we use a \textit{mixed integer semi-definite program} (MISDP) formulation to solve the problem of constructing a network topology and generating a pairwise distance between communicating robots.  The pairwise distances are constructed such that the communicating robots respect a minimum-distance collision constraint, but are within communication range. Our MISDP objective encourages communicating robots to spread out by maximizing the total pairwise distances between communicating robots.
	
	The pairwise distance output from the MISDP forms the input to the second stage of our approach, where we compute physical locations of the robot team.
	The communication topology and desired pairwise distances dictate an optimization objective and constraints on the physical locations of the robots.
	Optimizing the multi-robot formation is a highly nonconvex, possibly NP-hard problem. We employ a simulated annealing algorithm with penalty functions to escape bad local minima. It is noteworthy that, in our model, the centralized controller  is a monitor for the activities of the robot team. It intervenes to reposition robots only in the event of a robot resource failure. Also, sharing resources with other robots through the centralized controller (e.g., a star network) is unsuitable as, it would increase the communication cost and possibly saturate the communication bandwidth of the centralized controller.
	

	In the literature, the problem of resilience in multi-robot systems has been studied primarily in the context of constructing communication networks that preclude malicious robots in the network from exerting an adverse influence on the network~\cite{zhang2012robustness,Zhang2015}. Researchers have proposed notions of resilience in networks such as \textit{r-robustness}~\cite{LeBlanc2013} and \textit{p-fraction robustness}~\cite{LeBlanc2013} to characterize the resilience of the network in achieving consensus in the presence non-cooperative or malicious agents, in terms of network connectivity. In~\cite{Bonilla2017} algorithms for building resilient robot networks are introduced using this characterization of resilience. Analogous to our work, Hausman \textit{et al.}~\cite{hausman2015cooperative} also develops a notion for reasoning about the quality of networks neighboring to a given network. Since~\cite{hausman2015cooperative} focuses on target tracking applications, it compares sensing network topologies in terms of their ability to reduce the expected future uncertainty of the target position.
	
	In this work, we take a different approach to resilience by posing it as an {\em abstract} resource sharing problem in a heterogeneous robot team. By doing so, we make the following three contributions: 
	
	\begin{itemize}
		\item We pose the problem of resilience in a heterogeneous robot team in the context of inter-robot resource sharing. As resources become unavailable on robots due to failures, the system reconfigures to bring resources back into communication range as much as possible. We introduce the \textit{task inefficacy matrix} -- a description of on how far is the resource distribution in a configuration is from the maximal resource distribution -- and exploit it to drive network reconfiguration.
		\item We formally pose and solve the \hyperref[prob: configuration generation]{configuration generation problem} which generates a new communication topology and pairwise distances in response to a failure. 
		\item We give a solution to the~\hyperref[prob: formation synthesis]{formation synthesis problem} which generates appropriate values for robot positions satisfying the new configuration.
	\end{itemize}
	
	
	\section{Preliminaries and Notation}
	
	For any positive integer $z \in \Zp$, $[z]$ denotes the set $\{1,2, \cdots, z\}$. We use $\|\cdot\|$ to denote the standard Euclidean 2-norm and the induced 2-norm for vectors and matrices respectively. $\|\mathbf{M}\|_F$ denotes the Frobenius norm of a matrix $\mathbf{M} \in \R^{m_1 \times m_2}$, $\|\mathbf{M}\|_F \triangleq \sqrt{trace(\mathbf{M}^T\mathbf{M})}$. If we arrange the singular values of matrix $\mathbf{M}$ in nondecreasing order, then the $i^{th}$ singular value of $\mathbf{M}$ is denoted by $\sigma_i\{\mathbf{M}\}$. The nuclear norm (trace norm) of $\mathbf{M}$, defined as $\sum_{i=1}^{\min\{m_1, m_2\}} \sigma_i\{\mathbf{M}\}$, is denoted by $\|\mathbf{M}\|_*$. We use $\mathbf{1}^{m_1}$ and $\mathbf{1}^{m_1 \times m_2}$ to represent a vector and matrix of ones of appropriate dimensions, respectively. Similarly, $\mathbf{0}^{m_1}$ and $\mathbf{0}^{m_1 \times m_2}$ denote a vector and matrix of zeros respectively. For any vector $T \in \R^{m_1}$, $Diag(T)$ denotes a matrix with the elements of $T$ along its diagonal. Also, $diag(\mathbf{M})$ outputs a vector which contains the diagonal entries of matrix $\mathbf{M}$ as its elements. $[\mathbf{M}]_{i,j}$ denotes the $i,j$ entry of $\mathbf{M}$. Finally, $\mathcal{S}^m_+$ denotes the space of $m \times m$ symmetric positive semi-definite matrices. 
	A weighted undirected graph $\mathcal{G}$ is defined by the triplet $(\mathcal{V},\mathcal{E} \subseteq \mathcal{V} \times \mathcal{V}, \mathbf{A} \in \R^{|\mathcal{V}| \times |\mathcal{V}|}_{\geq 0})$, where $\mathbf{A}$ is the weighted adjacency matrix of the graph. Also, $\overline{\cE} = (\cV \times \cV) \setminus \cE$ denotes the edge complement of $\mathcal{G}$. The \textit{graph Laplacian matrix} $\mathbf{L}$ of $\mathcal{E}$ can be computed as
	\begin{align}
		\label{eqn:laplacian defin}
		\mathbf{L} = Diag(\mathbf{A}\cdot \mathbf{1}^{|\mathcal{V}|}) - \mathbf{A}.
	\end{align}
	We list a few properties of a graph Laplacian matrix used in this article~\cite{GodsilRoyle2001}:
	\begin{align}
		\label{eqn:lap property 1}
		\mathbf{L} \cdot \mathbf{1}^{|\mathcal{V}|} &= \mathbf{0}^{|\mathcal{V}|} \\
		\label{eqn:lap property 2}
		Trace(\mathbf{L}) &= \sum_{1\leq i < j \leq |\mathcal{V}|} [\mathbf{A}]_{i,j}.
	\end{align}

	\section{Problem Statement}
	\label{sec:problem_statement}

	We consider a team of $n$ heterogeneous robots labeled sequentially as $[n]$, equipped with different types of resources (e.g. sensors, memory, actuators, etc.). The team is assigned with a task of interest. For ease of identification, we denote the robot with label $i$ as $i_n$ . Also, let $X_i$ denotes the position vector $[x_i, y_i, z_i]^T \in \R^3$ of $i_n$ and $\{X_i\}_{i=1}^n$ denote the positions of the robots. A robot is equipped with at most $r$ distinct types of resources. 
	The set $[r]$ contains labels of the resources available within the heterogeneous team.
	
	We assert that all $r$ resources are essential to perform the assigned task. In other words, the team becomes incapable of performing the  task of interest if even a single resource among the $r$ resources is unavailable within the team. We assume that each robot can localize itself accurately in the environment. Further, we assume that the robots never lose their localization ability. 
	Finally, we assume that the communication graph (defined next) always has edge connectivity of at least 2, i.e., a single failure of a communication link during the task does not partition the communication graph.  
	
	We model the communication network using a dynamic undirected graph ${\mathcal{G} = (\mathcal{V}=[n],\ \mathcal{E} \subseteq \mathcal{V} \times \mathcal{V})}$, where the edges represent pairs of communicating robots. Based on the communication graph $\mathcal{G}$, we define its closed adjacency matrix $ \mathbf{\Bar{A}} $\cite{abbas2014characterizing} as follows:
	\begin{align}
		\label{eqn:adjacency closed matrix}
		[\mathbf{\Bar{A}}]_{i,j} = 
		\begin{cases}
			1 & \text{if}~~ (i,j) \in \cE~\text{or}~i = j\\
			0 & \text{otherwise.}
		\end{cases}
	\end{align}
	In addition, we define the neighborhood distance matrix $\mathbf{D}$, such that, 
	\begin{align}
		\label{eqn:neigh_dist_matrix}
		[\mathbf{D}]_{i,j} = 
		\begin{cases}
			\|X_i - X_j\| & \text{if}~~ (i,j) \in \cE~\text{or}~(j,i) \in \cE\\
			\quad \quad \infty & \text{otherwise.}
		\end{cases}
	\end{align}
	For brevity, we also use $d_{ij}$ to denote $[\mathbf{D}]_{i,j}$. Next we define a resilience metric which is a function of the communication network topology and the resource distribution in the robot team. As in~\cite{abbas2014characterizing},  we define a binary matrix $\{0,1\}^{n \times r}$,  which we refer to as the \textit{resource matrix} and denote by $\boldsymbol{\Gamma}_r$. The entries of $\boldsymbol{\Gamma}_r$ are computed as follows, 
	\begin{align}
		\label{eqn:resource matrix}
		[\boldsymbol{\Gamma}_r]_{i,j} =  
		\begin{cases}
			1 & i_n~ \text{has resource}~j  \in [r] \\
			0 & \text{otherwise.}
		\end{cases}
	\end{align}
	We term the tuple $(\mathcal{G}, \mathbf{D}, \boldsymbol{\Gamma}_r)$ as a \textit{configuration} of the heterogeneous multi-robot team and denote it by $\mathcal{C}$.
	We define the \textit{task inefficacy matrix} $\mathbf{V}$ of a configuration $\mathcal{C}$ as
	\begin{align}
		\label{eqn:inefficacy matrix}
		\mathbf{V(\Bar{A}, \boldsymbol{\Gamma})} = n \cdot \mathbf{1}^{n \times r} - \mathbf{\Bar{A}} \boldsymbol{\Gamma}_r.
	\end{align}
	Finally, we define the key quantity used to measure the inability of a heterogeneous multi-robot team with configuration $\mathcal{C}$ to perform the task - the \textit{task inefficacy} of a heterogeneous multi-robot team. The task inefficacy of a heterogeneous multi-robot team with configuration $\mathcal{C}$ is computed as
	\begin{align}
		\label{eqn: task inefficacy}
		\|\mathbf{V(\Bar{A}, \boldsymbol{\Gamma})}\|_* = \|n \cdot \mathbf{1}^{n \times r} - \mathbf{\Bar{A}} \boldsymbol{\Gamma}_r\|_*.
	\end{align}
	Next, we examine $\mathbf{V}$  to understand why $\|\mathbf{V}\|_*$ is a reasonable metric to quantify the task inefficacy of $\mathcal{C}$.  The product matrix $\mathbf{\Bar{A}}\cdot \boldsymbol{\Gamma}_r$ encodes information regarding the distribution of resources within the heterogeneous multi-robot team. Particularly,  $[\mathbf{\Bar{A}}\cdot \boldsymbol{\Gamma}_r]_{i,j}$ describes the exact number of robots in the neighborhood of robot $i_n$, including itself, with resource $j$. This matrix becomes $n \cdot \mathbf{1}^{n \times r}$ if each robot is equipped with every resource and the communication graph is fully connected, resulting in the most task-efficient configuration. The task inefficacy matrix thus describes how far a configuration is from the maximal configuration.
	Therefore,
	we minimize a metric which measures the distance between these matrices with a suitable norm. The nuclear norm is a surrogate measure to quantify the rank of a matrix. It is used extensively in the compressed sensing literature to compute low rank matrices~\cite{recht2010guaranteed}.  As the robot configuration approaches the maximal configuration the \textit{task inefficacy matrix} becomes sparser.  We use the nuclear norm as a heuristic to measure the sparseness of the task inefficacy matrix. Although in general, low rank matrices need not be sparse, we found that the \textit{nuclear norm} of the task inefficacy matrix works well for solving our problem and we employ it here.  We may explore other matrix norms in future work. When it improves clarity, we will also refer to maximizing the \emph{task efficacy} instead of minimizing the task inefficacy.
	
	We refer to a resource matrix $\boldsymbol{\Gamma}_r$ as a \textit{feasible resource matrix}, if $\boldsymbol{\Gamma}_r^T\mathbf{1}^n > \mathbf{1}^r$, where $>$ is applied elementwise. We term a resource matrix where $\boldsymbol{\Gamma}_r^T\mathbf{1}^n \ngtr \mathbf{1}^r$ as \emph{infeasible}.
	Within our framework, any heterogeneous robot team having an infeasible resource matrix lacks the ability to perform the overall task. In this paper, a \textit{resource failure} is a function that consumes a resource matrix (excluding the $n \times r$ matrix of zeros) and generates a resource matrix by setting a random nonzero entry of the consumed resource matrix to zero. A \textit{tolerable resource failure} maps a feasible resource matrix to another feasible resource matrix, while a \textit{catastrophic resource failure} maps a feasible resource matrix to an infeasible resource matrix.
	Since a catastrophic resource failure renders the team incapable of performing the task, we consider resilience only in the context of tolerable resource failures.
	
	Consider a sequence of countably infinite resource failures $\mathcal{F} = [f_1, f_2, \cdots f_{\infty}]$ acting on a feasible resource matrix $\boldsymbol{\Gamma}_r$ sequentially. Now, let $\mathcal{F}_{n_f}=[f_1, f_2, \cdots, f_k, \cdots, f_{n_f}]$ be the first $n_f \in \Zp$ tolerable resource failures in $\mathcal{F}$, such that $f_{n_f + 1} \in \mathcal{F}$ is a catastrophic failure. The $k^{th}$ tolerable resource failure in $\mathcal{F}_{n_f}$ is denoted as $f_k$. We denote $\boldsymbol{\Gamma}_r[k]$ as the resultant resource matrix after first $k$ tolerable resource failures in $\mathcal{F}_{n_f}$ acted on $\boldsymbol{\Gamma}_r$. $\boldsymbol{\Gamma}_r[k]$ can be recursively defined as $\boldsymbol{\Gamma}_r[k] = f_k(\boldsymbol{\Gamma}_r[k-1]),\ \boldsymbol{\Gamma}_r[0] = \boldsymbol{\Gamma}_r$. 
	We term $\mathcal{C}[k-1]$ as the configuration of the heterogeneous multi-robot team before $f_k$ occurred. 
	
	We formally define the problems addressed in this paper:
	
	\begin{problem}
		\label{prob: configuration generation}
		\textbf{Configuration generation:} Given a tolerable resource failure $f_k$, an associated feasible resource matrix $\boldsymbol{\Gamma}_r[k]$ and a heterogeneous multi-robot team configuration $\mathcal{C}[k-1]$ find a new configuration $\mathcal{C}[k]$ such that,
		\begin{enumerate}
			\item $\mathcal{G}[k]$ is a connected graph,
			\item $\| n \cdot \mathbf{1}^{n \times r} - \mathbf{\Bar{A}}[k] \boldsymbol{\Gamma}_r[k] \|_* < \| n \cdot \mathbf{1}^{n \times r} - \mathbf{\Bar{A}}[k-1] \boldsymbol{\Gamma}_r[k] \|_*$
			\item $\|\mathbf{\Bar{A}}[k] -  \mathbf{\Bar{A}}[k-1]\|_F^2 \leq ne$,
			where $ne \in \Zp$ is a user-chosen parameter specifying the number of edges to be modified in $\bar A[k-1]$ to produce $\bar A[k]$
			and 
			\item the sum of distances between communicating robots is maximized.
		\end{enumerate}
	\end{problem}
	
	\begin{problem}
		\label{prob: formation synthesis}
		\textbf{Formation synthesis:} Given a heterogeneous multi-robot team configuration $\mathcal{C}[k]$, generate coordinates $X_1[k], \dots, X_n[k] \in \R^3$ for the robots that best realize the given configuration. We describe this problem more precisely in~\autoref{subsec:Formation synthesis}.
	\end{problem}
	
	The condition of graph connectivity in \hyperref[prob: configuration generation]{Problem~\ref{prob: configuration generation}} is required to ensure cooperative task performance of the heterogeneous multi-robot team. The second condition in \hyperref[prob: configuration generation]{Problem~\ref{prob: configuration generation}} states that the new configuration should improve the task efficacy of the multi-robot system. The third condition ensures that graph topologies $\mathcal{G}[k-1]$ and $\mathcal{G}[k]$ differ from each other in at most $ne$ edges. Although the final condition does not aid in resilience, it is designed to cause robots to spread out in space (this useful \eg for coverage).
	The following section describes our solution procedure for these two problems. 
	
	\section{Procedure}  
	\label{sec:Procedure}
	
	\begin{figure*}[!t]
		\centering
		\hspace{3mm}
		\includegraphics[width=0.75\linewidth, height=.3\linewidth]{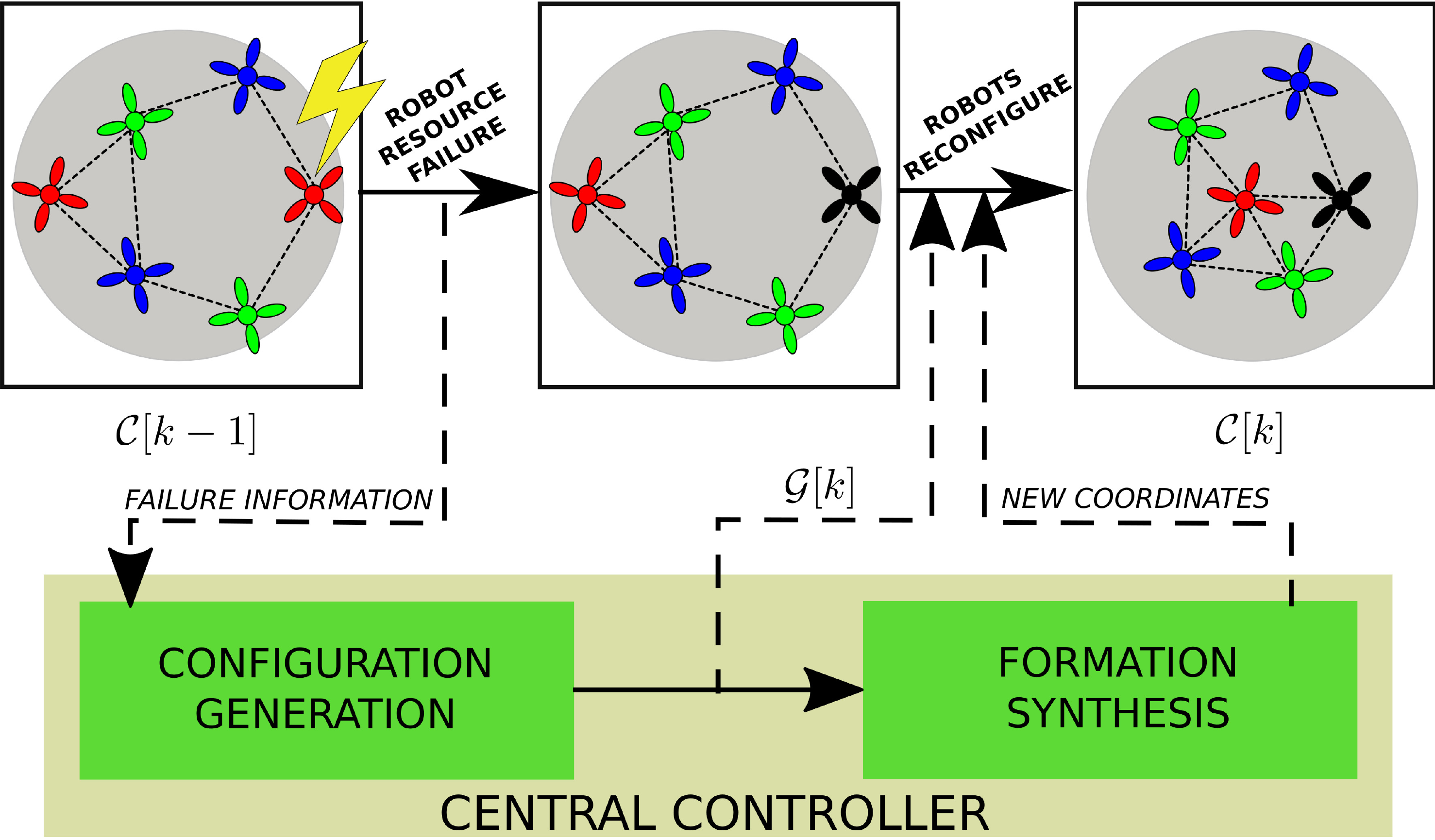}
		\caption{Basic outline of our approach. When a resource is lost, \emph{configuration generation} selects edges to modify the communication graph. Then, \emph{formation synthesis} assigns robots to physical locations that support the desired graph topology. }
		\label{fig:schematic}       
		\vspace{-5mm}
	\end{figure*}
	
	We explain our methods for the solving \hyperref[prob: configuration generation]{Problem~\ref{prob: configuration generation}} and \hyperref[prob: formation synthesis]{Problem~\ref{prob: formation synthesis}} in this section.
	When a resource failure occurs, information about the failure is transmitted to a central controller. The central controller checks if the resource failure is tolerable or catastrophic. If catastrophic, it commands the robots to return to their base station as the team no longer retains the necessary resources to perform the assigned task.  If the resource failure is tolerable, the central controller uses a two-step procedure to generate a new configuration and an associated set of robot coordinates which realizes the generated configuration in three-dimensional space. \autoref{fig:schematic} gives a schematic illustration of our multi-robot reconfiguration strategy in the event of a tolerable resource failure. We refer to these steps as \textit{configuration generation} and \textit{formation synthesis}. Specifically, configuration generation and formation synthesis steps are solutions to  \hyperref[prob: configuration generation]{Problem~\ref{prob: configuration generation}} and \hyperref[prob: formation synthesis]{Problem~\ref{prob: formation synthesis}} respectively. The following subsections describe each step in further detail.
	
	\subsection{Configuration generation}
	\label{subsec:config gen}
	
	We now describe our proposed solution to \hyperref[prob: configuration generation]{Problem~\ref{prob: configuration generation}}. We reformulate the problem as a mixed integer semi-definite program (MISDP) and then solve reformulated MISDP. MISDP formulations are extensively used by researchers for graph topology design problems~\cite{rafiee2010optimal,xue2014distributed}.  We first state our MISDP mathematically; in the following discussion we argue that it captures the facets of \hyperref[prob: configuration generation]{Problem~\ref{prob: configuration generation}}. A MISDP formulation of \hyperref[prob: configuration generation]{Problem~\ref{prob: configuration generation}} can be written as 
	\begin{align}
		\label{eqn: MISDP obj}
		\minimize_{\substack{\mathbf{L} \in \mathcal{S}^n_+,\ \mu \in \R_{> 0}, \\ \boldsymbol{\Pi} \in \{0,1\}^{n\times n}}} \quad &  \text{Trace}(\mathbf{L}) \\
		\label{eqn:cons:zero sum}
		\text{subject to} ~~ & \mathbf{L}\cdot\mathbf{1}^n = \mathbf{0}^n\\
		\label{eqn:cons:connectivity}
		~~ &\frac{1}{n} \mathbf{1}^T \mathbf{1}  + \mathbf{L} \succeq \mu \mathbf{I} \\
		\label{eqn:cons:diag binary}
		~~ & diag(\boldsymbol{\Pi}) = \mathbf{1}^n\ \forall\ i \in [n]\\
		\label{eqn:cons:binary sym}
		~~ & \boldsymbol{\Pi} = \boldsymbol{\Pi}^T\\
		\label{eqn:cons:Lap diag}
		~~ & [\mathbf{L}]_{i,i} \geq c_{max} \forall\ i \in [n]\\
		\label{eqn:cons:Lap off diag min}
		~~ & [\mathbf{L}]_{i,j} \geq c_{min} \boldsymbol{\Pi}_{i,j}  \forall~(i, j) \in [n]^2,~i \neq j \\
		\label{eqn:cons:Lap off diag max}
		~~ & [\mathbf{L}]_{i,j} \leq c_{max} \boldsymbol{\Pi}_{i,j} \forall ~(i, j) \in [n]^2,~i \neq j\\
		\label{eqn:cons:topology near}
		~~ & \|\boldsymbol{\Pi} - \mathbf{\Bar{A}}[k-1] \|_F^2 \leq ne \\
		\label{eqn:cons:reduce inefficacy}
		~~&\|\mathbf{V(\boldsymbol{\Pi}, \boldsymbol{\Gamma}_r)}\|_* < \|\mathbf{V(\mathbf{\Bar{A}}[k-1], \boldsymbol{\Gamma}_r)}\|_*.
	\end{align}
	The decision variable $\mathbf{L}$ is a weighted Laplacian of a graph that has the same topology as $\boldsymbol{\Pi}$ and $\Bar{\mathbf{A}}[k]$, 
	and whose weights will be used to compute the desired neighborhood distance matrix $\mathbf{D}[k]$.
	${0<c_{min} < |\mathbf{L}_{i,j}| < c_{max}}$ gives the range of nonzero values of the off-diagonal entries of $\mathbf{L}$.
	\hyperref[eqn:cons:zero sum]{Constraint~\ref{eqn:cons:zero sum}} encodes the property of a graph Laplacian described in \autoref{eqn:lap property 1}.
	It is known~\cite[Proposition 1]{sundin2017connectedness} that any undirected graph is connected if and only if $\mathbf{L} + \frac{1}{n} \mathbf{1}^T \mathbf{1}$ is a positive-definite matrix,
	therefore \hyperref[eqn:cons:connectivity]{Constraint~\ref{eqn:cons:connectivity}} ensures that the graph represented by $\mathbf{L}$ must be connected. 
	\hyperref[eqn:cons:diag binary]{Constraint~\ref{eqn:cons:diag binary}} and \hyperref[eqn:cons:binary sym]{Constraint~\ref{eqn:cons:binary sym}}   models $\boldsymbol{\Pi}$ as the closed adjacency matrix of a graph. \hyperref[eqn:cons:Lap off diag max]{Constraint~\ref{eqn:cons:Lap off diag max}} and \hyperref[eqn:cons:Lap off diag min]{Constraint~\ref{eqn:cons:Lap off diag min}} force $\mathbf{L}$ and $\boldsymbol{\Pi}$ to model the Laplacian matrix and closed adjacency matrix of graphs with same topology, respectively. In other words, the entries of $\mathbf{L}$ are nonzero if and only if the corresponding entries of $\boldsymbol{\Pi}$ are nonzero. 
	\hyperref[eqn:cons:topology near]{Constraint~\ref{eqn:cons:topology near}} and \hyperref[eqn:cons:reduce inefficacy]{Constraint~\ref{eqn:cons:reduce inefficacy}}  are a direct consequence of the conditions in \hyperref[prob: configuration generation]{Problem~\ref{prob: configuration generation}}
	
	The optimization problem described in \autoref{eqn: MISDP obj}-\autoref{eqn:cons:reduce inefficacy} is a MISDP as it involves an integer matrix ($\boldsymbol{\Pi}$) and a semi-definite matrix ($\mathbf{L}$) as decision variables. Unlike the integer matrix (binary matrix) $\boldsymbol{\Pi}$, whose sole purpose is modeling the closed adjacency matrix of the underlying graph, the Laplacian matrix $\mathbf{L}$ serves two purposes. First, it gives a simple way  to incorporate the connectivity constraint of the associated graph in the form of a \textit{linear matrix inequality}~\cite{boyd1994linear}. Second, we use the off-diagonal entries of $\mathbf{L}$ to compute the neighborhood distance matrix $\mathbf{D}[k]$ associated with the new configuration $\mathcal{C}[k]$. The computation of neighborhood distance matrix $\mathbf{D}[k]$ based on a Laplacian matrix $\mathbf{L}$ is performed as follows:
	\begin{equation}
	\label{eqn:dist matrix computation}
	[\mathbf{D}[k]]_{i,j} = 
	\begin{cases}
	\kappa(|[\mathbf{L}]_{i,j}|- c_{min}) + d_{mc} & \text{if}~~[\mathbf{L}]_{i,j} < 0\\
	\quad \quad \infty & \text{otherwise},
	\end{cases}
	\end{equation}
	where $\kappa$ (a constant defined as $\kappa \coloneqq \frac{d_{s} - d_{mc}}{c_{max}-c_{min}}$, $d_s \in R_{> 0}$) is the minimum safe distance between robots and ${d_{mc} \in \R_{> 0}}$ is the minimum distance for non-communicating pairs, such that $d_{mc} > d_s$. From \autoref{eqn:dist matrix computation} it is apparent that the distances between communicating pairs of robots increase as the absolute value of the corresponding Laplacian matrix entries decrease. As a result of \autoref{eqn:lap property 2}, maximizing the sum of inter-robot distances between communicating robots can be achieved by minimizing the trace of $\mathbf{L}$.  Finally, the constraint described in \autoref{eqn:cons:Lap diag} implies that robots with fewer neighbors should be close to its neighbors. This enables the robots with weak connectivity to remain connected with the communication network, even in presence of small external disturbances (e.g. wind).  Thus, solving the MISDP described in this section in turn solves \hyperref[prob: configuration generation]{Problem~\ref{prob: configuration generation}}, thereby generating a new configuration $\mathcal{C}[k]=(\mathcal{G}(\mathbf{\Bar{A}[k]}=\boldsymbol{\Pi})[k],\mathbf{D}[k],\boldsymbol{\Gamma}_r[k])$ for the team. The neighborhood distance matrix $\mathbf{D}[k]$ forms the input to the formation synthesis stage (along with the current formation coordinates to serve as an initial guess in the optimization procedure).


	\subsection{Formation synthesis}
	\label{subsec:Formation synthesis}
	In this subsection, we introduce a procedure to assign a physical location to each robot
	based on the desired distances $\bD$ specified from the procedure of~\autoref{subsec:config gen}.
	We drop the resource failure index $[k]$ for brevity.
	The primary objective is to render $\{X_i\}_{i=1}^n$ such that they match desired inter-robot distances obtained from the previous step.
	
	It is possible that the desired $d_{i,j}$ impose a geometrically impossible set of pairwise distances.
	To account for this possibility, we introduce the constraint that each pairwise distance must exceed $d_{mc}$ to ensure that no two robots collide.
	Additionally, to demonstrate our method in an indoor experimental space, a bounding volume constraint on the whole formation is needed.
	These considerations produce a constrained optimization problem:
	
	\begin{align}
		\label{obj}
		\minimize \quad & \omit\rlap{$\displaystyle\sum_{(i,j) \in \cE} (\|X_i - X_j\|_2 - d_{ij})^2$} \\
		\label{con-collision}
		\text{subject to} \quad & d_s \leq \|X_i - X_j\|_2 \ \leq d_{mc} & & \forall\ (i, j) \in \cE \\
		\label{con-nonneighbor}
		& d_{mc} \leq \|X_i - X_j\|_2 & & \forall\ (i, j) \in \overline{\cE} \\
		\label{con-box}
		& B^{\min} \leq X_i \leq B^{\max} & & \forall\ i \in V,
	\end{align}
	where $B^{\min},\ B^{\max} \in \R^3$ are the minimum and maximum extents of an axis-aligned bounding box, with the operator $\leq$ applied elementwise in~\ref{con-box}.
	
	The unconstrained version of problem~\ref{obj} is similar to the \emph{graph drawing}~\cite{zheng-graph-drawing-sgd} and \emph{multidimensional scaling}~\cite{klimenta2012graph} problems.
	Graph-drawing variants with constraints similar to~\ref{con-collision}--\ref{con-box}
	are considered in~\cite{davidson-drawing-graphs-SA,dwyer-simple-constrained}.
	In fact, the feasibility problem composed ]only of the constraints~\ref{con-collision} and~\ref{con-box} is a type of \emph{sphere-packing} problem,
	a class of problems recognized as highly difficult for centuries.
	The two-dimensional version was shown to be NP-complete by~\cite{demaine-circle-packing-nphard},
	while some other packing variants are not even established as members of NP.
	Some interesting reviews of packing problems are found in~\cite{hifi-review-circle-sphere-packing}.
	
	Consequently we cannot hope to optimize~\ref{obj}-\ref{con-box} exactly, but in practice, randomized algorithms perform well.
	Here, we use a simulated annealing approach with constraints approximated by penalty functions.
	\subsubsection{Simulated annealing}
	Simulated annealing (SA) is a well-known stochastic global optimization algorithm
	that can escape poor local minima by occasionally taking random steps that cause an \emph{increase} in the minimization objective.
	We propose to use SA for the optimization problem~\ref{obj}
	because the set of feasible solutions to sphere-packing problems do not form connected sets in general.
	Thus, strictly local optimization methods are not appropriate.
	In this section we briefly recap the algorithmic framework and specify our objective and algorithm hyperparameters for applying SA to solve~\ref{obj}.
	For additional details on SA refer  to~\cite{kirkpatrick-SA,henderson-SA}.
	
	\begin{algorithm}[t]
		
		\begin{algorithmic}
			\Procedure{Simulated-Annealing}{$x_0$ : initial guess}
			\State $x \gets x_0$
			\Repeat
			\State $x' \gets \Call{Propose}{x}$
			\If{$E(x') < E(x)$} $x \gets x'$
			\Else $\ x \gets x'$ w/ probability $\exp(-T(E(x') - E(x)))$
			\EndIf
			\State $T \gets \Call{Cooling}{T}$
			\Until{stopping criterion met}
			\EndProcedure
			\Procedure{$x' =$ Propose}{x}
			\State sample $j \sim \text{Uniform}([n]),\ d \sim \text{Uniform}([3])$
			\State sample $\delta \sim \text{Uniform}([-\delta_{\max},\ \delta_{\max}])$
			\State $x' = x$
			\State $x'_{j,d} \gets x'_{j,d} + \delta$
			\EndProcedure
		\end{algorithmic}
		\caption{Simulated annealing (\cite{kirkpatrick-SA,henderson-SA})}
		\label{algSA}
		
	\end{algorithm}
	
	We recap the algorithmic framework of SA in Algorithm~\ref{algSA}.
	$\Call{Propose}{x}$ generates a new value $x'$ in the neighborhood of $x$,
	$E(x)$ is the objective or \emph{energy} function,
	and $T$ is a \emph{temperature} controlling the probability of taking a step that increases the minimization objective,
	with $\Call{Cooling}{T}$ implementing a cooling schedule such that $T$ decreases over time.
	The implementation of $\Call{Propose}{x}$ in Algorithm~\ref{algSA}
	is a generic one applicable to many problems with continuous decision variables.
	We use exponential cooling ($\Call{Cooling}{T} = \gamma T$ for $0 \ll \gamma < 1$),
	and our stopping criterion is a fixed number of iterations.
	These are straightforward choices that yielded good solutions in our experiments.
	
	\begin{figure*}[ht]
		\vspace{2mm}
		\begin{tabular}{|c|c|c|c|c|}
	\centering	
	\subcaptionbox{Before $1^{st}$ failure  \label{subfig:scr_t_0}}{\includegraphics[width=.175\linewidth, height=.14\linewidth]{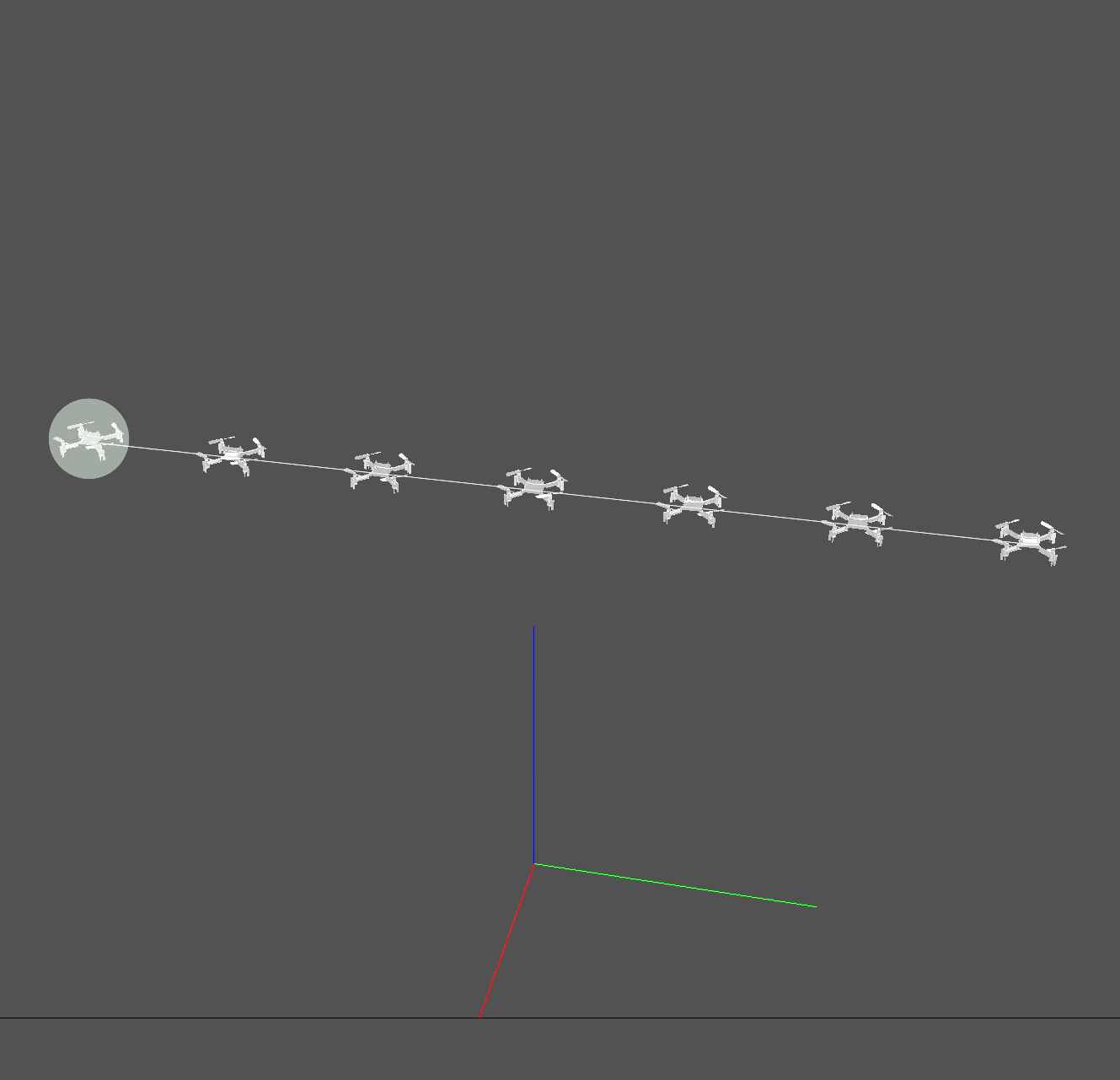}}
	&
	\subcaptionbox{Before $6^{th}$ failure  \label{subfig:scr_t_1}}{\includegraphics[width=.175\linewidth, height=.14\linewidth]{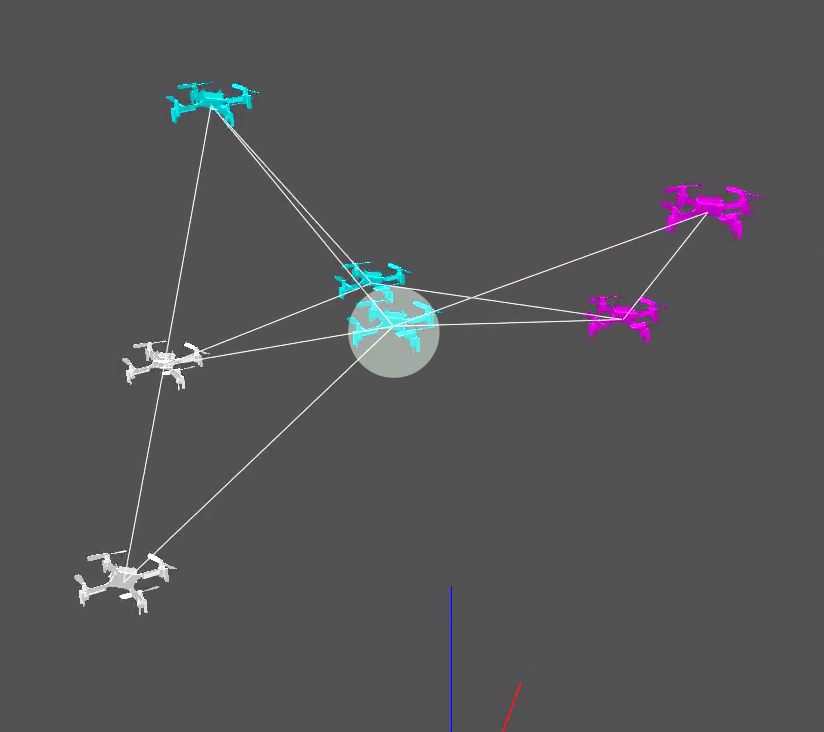}} 
	&		
	\subcaptionbox{Before $12^{th}$ failure  \label{subfig:scr_t_2}}{\includegraphics[width=.175\linewidth, height=.14\linewidth]{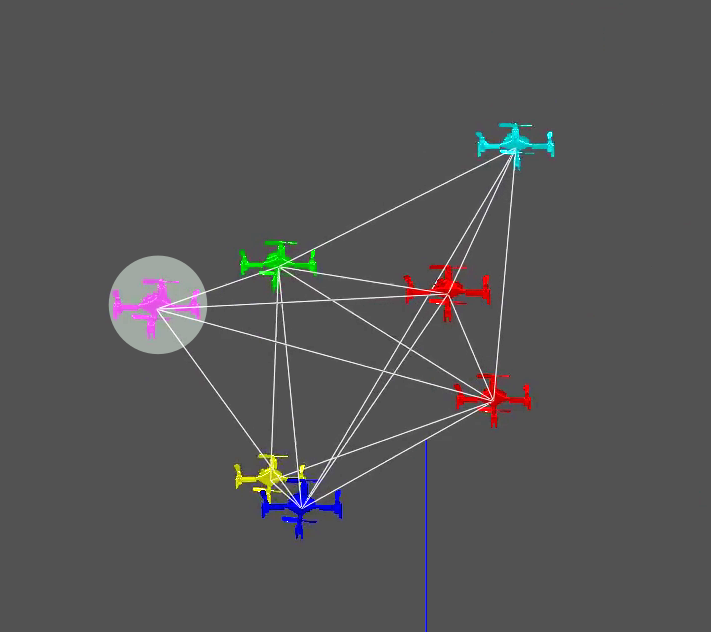}}
	&
	\subcaptionbox{Before $14^{th}$ failure \label{subfig:scr_t_3}}{\includegraphics[width=.175\linewidth, height=.14\linewidth]{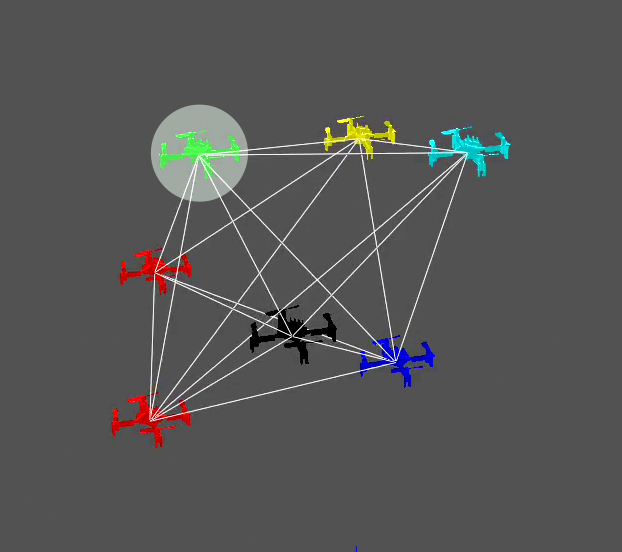}}
	&
	\subcaptionbox{Before final failure \label{subfig:scr_t_4}}{\includegraphics[width=.175\linewidth, height=.14\linewidth]{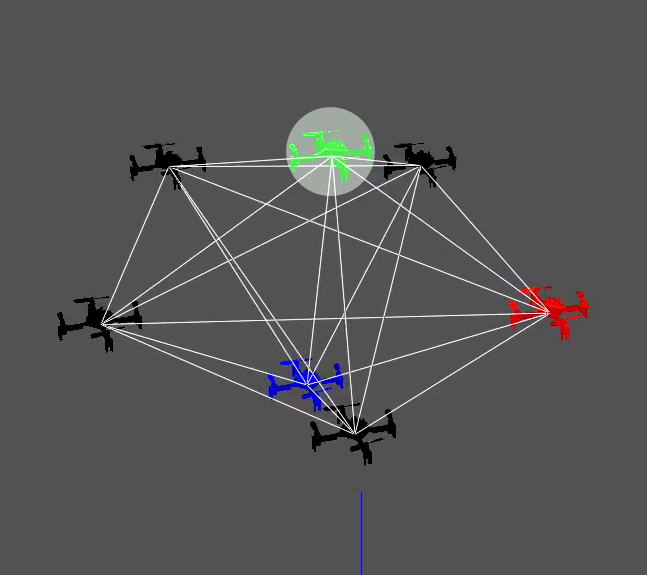}} 
	\\
	\subcaptionbox{After $1^{st}$ failure \label{subfig:scr_t_0_}}{\includegraphics[width=.175\linewidth, height=.14\linewidth]{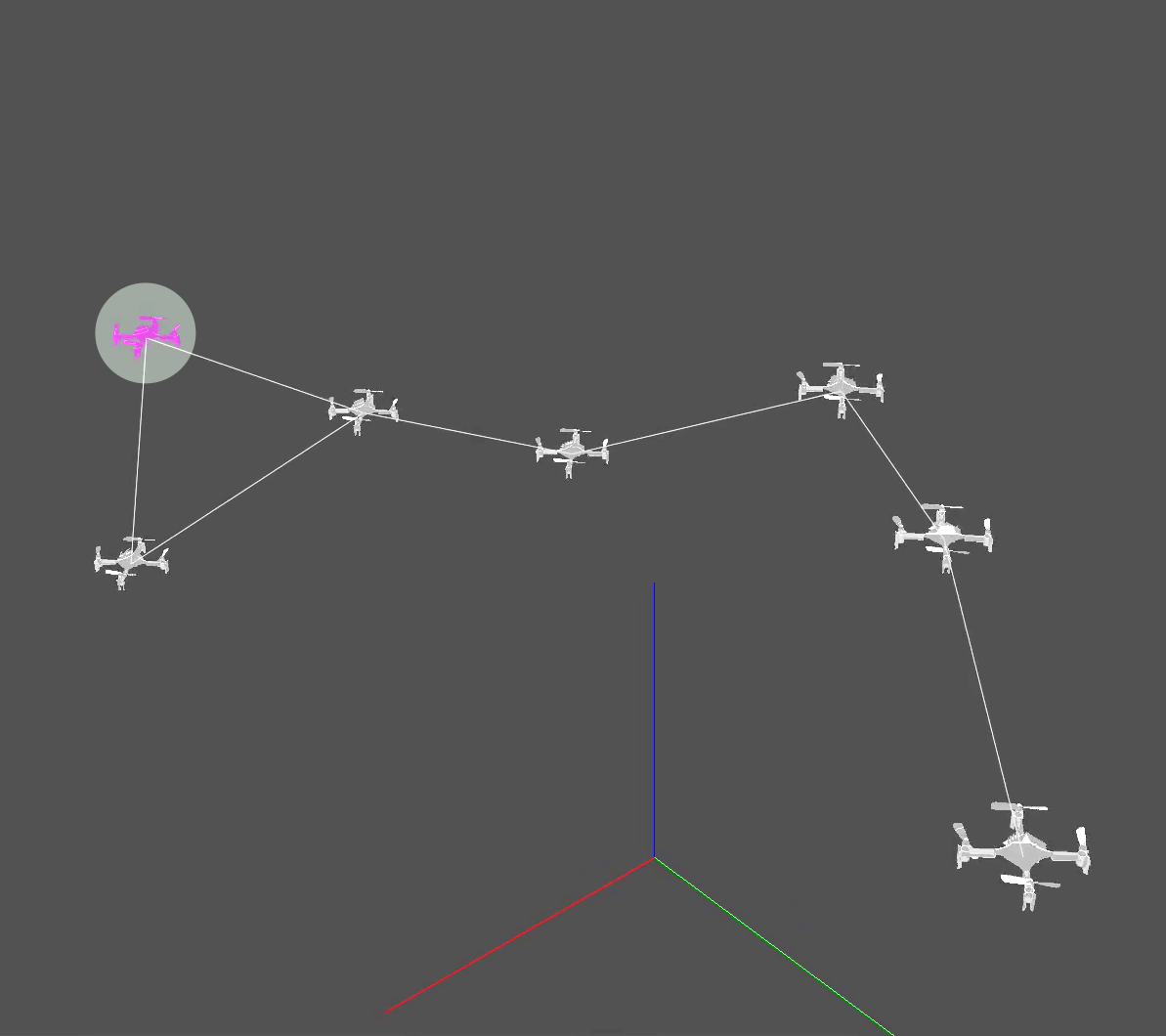}}
	&
	\subcaptionbox{After $6^{th}$ failure \label{subfig:scr_t_1_}}{\includegraphics[width=.175\linewidth, height=.14\linewidth]{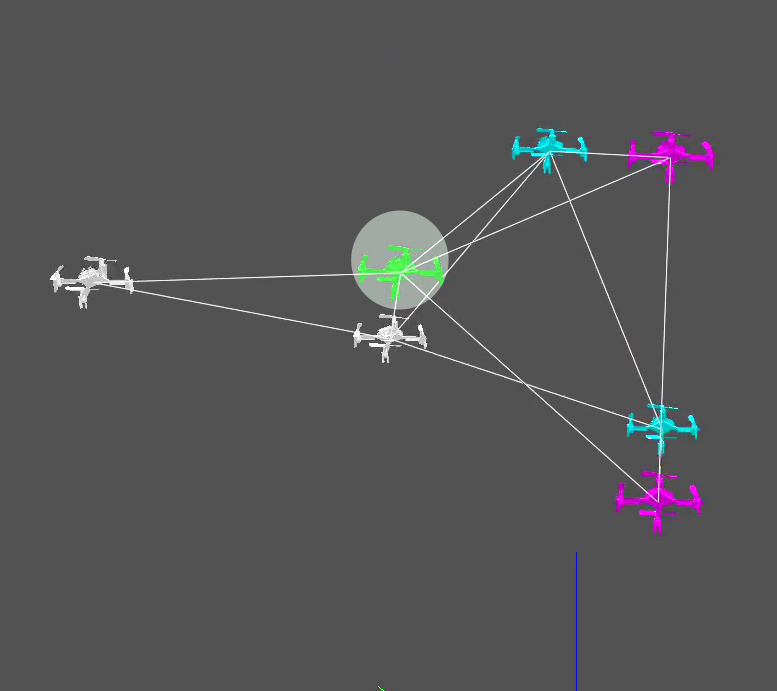}} 
	&		
	\subcaptionbox{ After $12^{th}$ failure \label{subfig:scr_t_2_}}{\includegraphics[width=.175\linewidth, height=.14\linewidth]{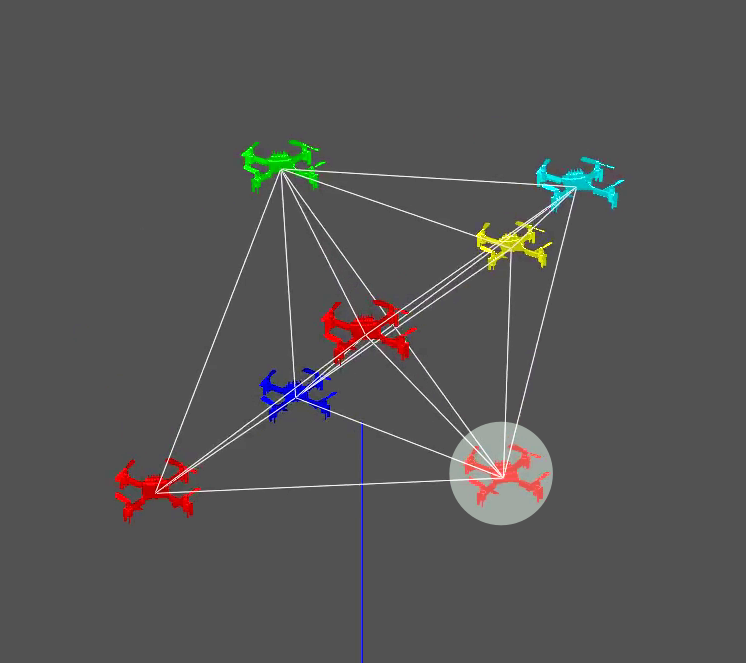}}
	&
	\subcaptionbox{After $14^{th}$ failure \label{subfig:scr_t_3_}}{\includegraphics[width=.175\linewidth, height=.14\linewidth]{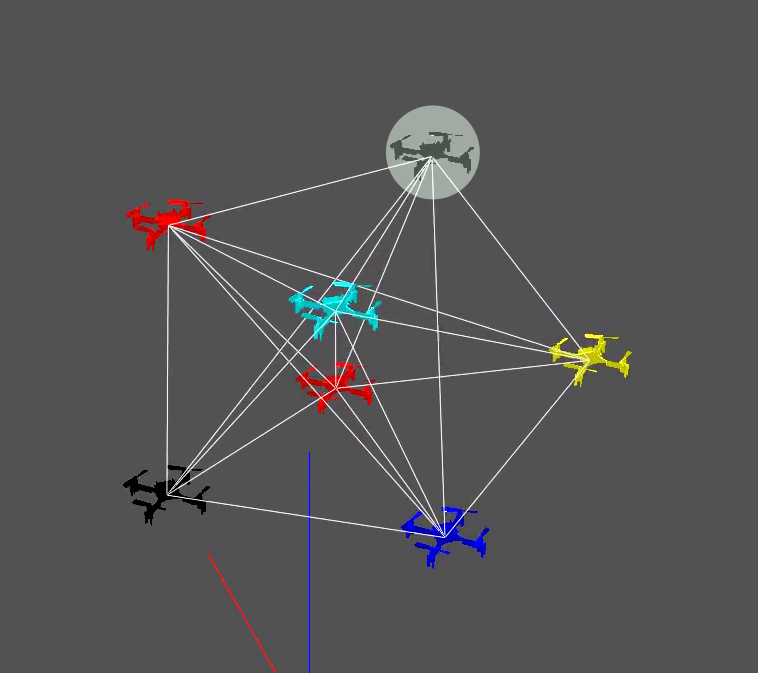}}
	&
	\subcaptionbox{ After final failure \label{subfig:scr_t_4_}}{\includegraphics[width=.175\linewidth, height=.14\linewidth]{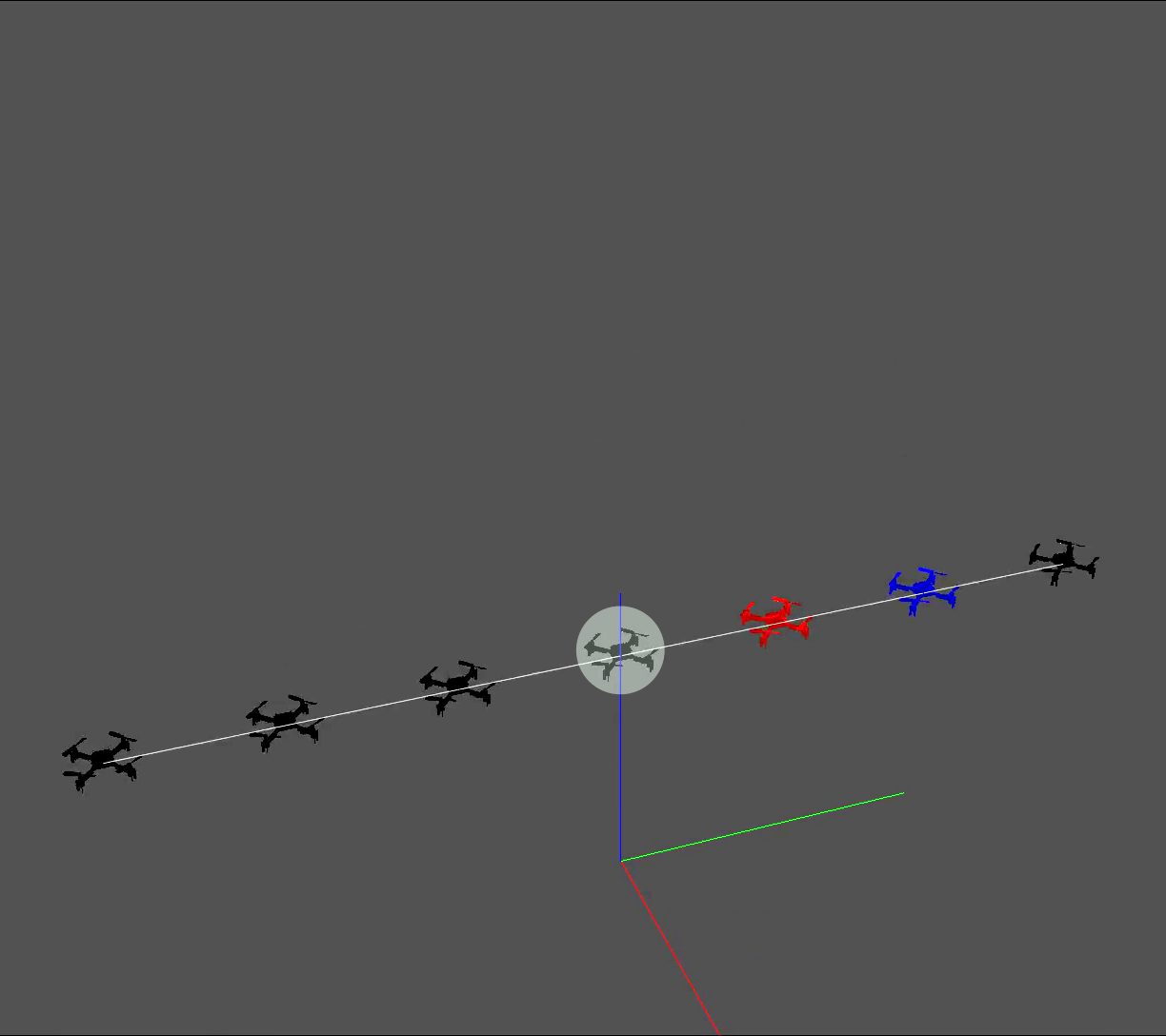}} 
\end{tabular}
\caption{Screenshots of a simulation in which seven quadrotors are changing their formation in the event of resource failures. The simulation was visualized using the visualizing tool in our Crazyswarm simulation architecture \cite{hoenig-preiss-planning}. The figures in the top row depicts the formation of the quadrotors before the occurrence of a resource failure. The corresponding figures on the bottom portray the formation after 1) the resource failure occurs and is detected, 2) a new communication edge is chosen, 3) the robots move to their new formation. The quadrotor whose resource fails is enveloped using a filled white circle.  }  
\label{fig:sim screenshots}
		\vspace{-5mm}
	\end{figure*}
	
	We capture the constraints in the energy function $E(x)$ by adding penalty functions:
	\begin{gather}
		\label{eqn:SA-energy}
		E(x) = 
		\sum_{(i,j) \in \cE} (\|X_i - X_j\|_2 - d_{ij})^2
		+ \sum_{(i,j) \in \overline{\cE}} p_H(d_{mc} - \|X_i - X_j\|_2) \notag \\
		+ \sum_{(i,j) \in \cE} \left[
		p_H(d_s - \|X_i - X_j\|_2)
		+ p_H(\|X_i - X_j\|_2 - d_{mc})
		\right] \notag \\
		+ \sum_{i \in V} \sum_{d \in [3]}
		p_H(X_{i,d} - B^{\max}_d)
		+ p_H(B^{\min}_d - X_{i,d}),
	\end{gather}
	where $p_H$ denotes a penalty function satisfying the property
	\begin{equation}
	\lim_{H \to \infty}\ p_H(y) =
	\begin{cases}
	\infty &: y > 0 \\
	0 &: y < 0.
	\end{cases}
	\end{equation}
	The ``hardness'' parameter $H > 0$ increases over iterations, analogous to the decay of $T$.
	Here we use the exponential penalty $p_H(y) = e^{Hy}$.
	While the literature on penalty function methods typically includes the property that $p_H(y) = 0$ within the feasible set~\cite{luenberger-ye-linprog},
	for example  $p_H(y) = \max\{0, T y^3\}$, 
	we observed solutions with values of~\ref{obj} closer to optimal using an exponential penalty. We leave further investigation into this for the future.
	We emphasize that the method of penalty functions does not guarantee that the solution will be feasible
	according to the hard constraints~\ref{con-collision}--\ref{con-box}.
	It is therefore necessary to perform a final feasibility check on the SA output after termination,
	and possibly restart or fall back on a more computationally expensive exact algorithm.
	In our experiments, we run SA for 20,000 steps.
	We choose $\gamma$ such that $T$ decays from $1$ to $10^{-8}$, and the growth constant for $H$ such that $H$ increases from $1$ to $10^3$.
	We let $\delta_{\max} = d_s / 10$. Note that tunings of $T$ and $H$ are sensitive to the overall scale of the distances $\mathbf{D},\ d_s,\ d_{mc}$ involved in the problem ($\approx \SI{1}{\meter}$ in our experiments).
	
	\subsection{Formation change motion planning}
	To physically realize the formation change induced by a resource failure, we must ensure that the robots can move between formations without collisions.
	We employ the method of~\cite{hoenig-preiss-planning} to plan a set of piecewise polynomial trajectories
	that transition the robots from configuration $X[k]$ to $X[k+1]$ safely.
	This method is appropriate for quadrotors in three-dimensional environments with obstacles.
	We emphasize that other multi-robot motion planning algorithms can be used as needed depending on the type of robot, the task, environment map representation, and so on.

	\section{Simulation Results}
	\label{sec:simulation}
	
	
	\begin{figure*}[ht]
		\centering
		\begin{tabular}{ccc}
			\subcaptionbox{$p_r=20\%$   \label{subfig:valid 20}}{\includegraphics[width=.3\linewidth, height=.16\linewidth]{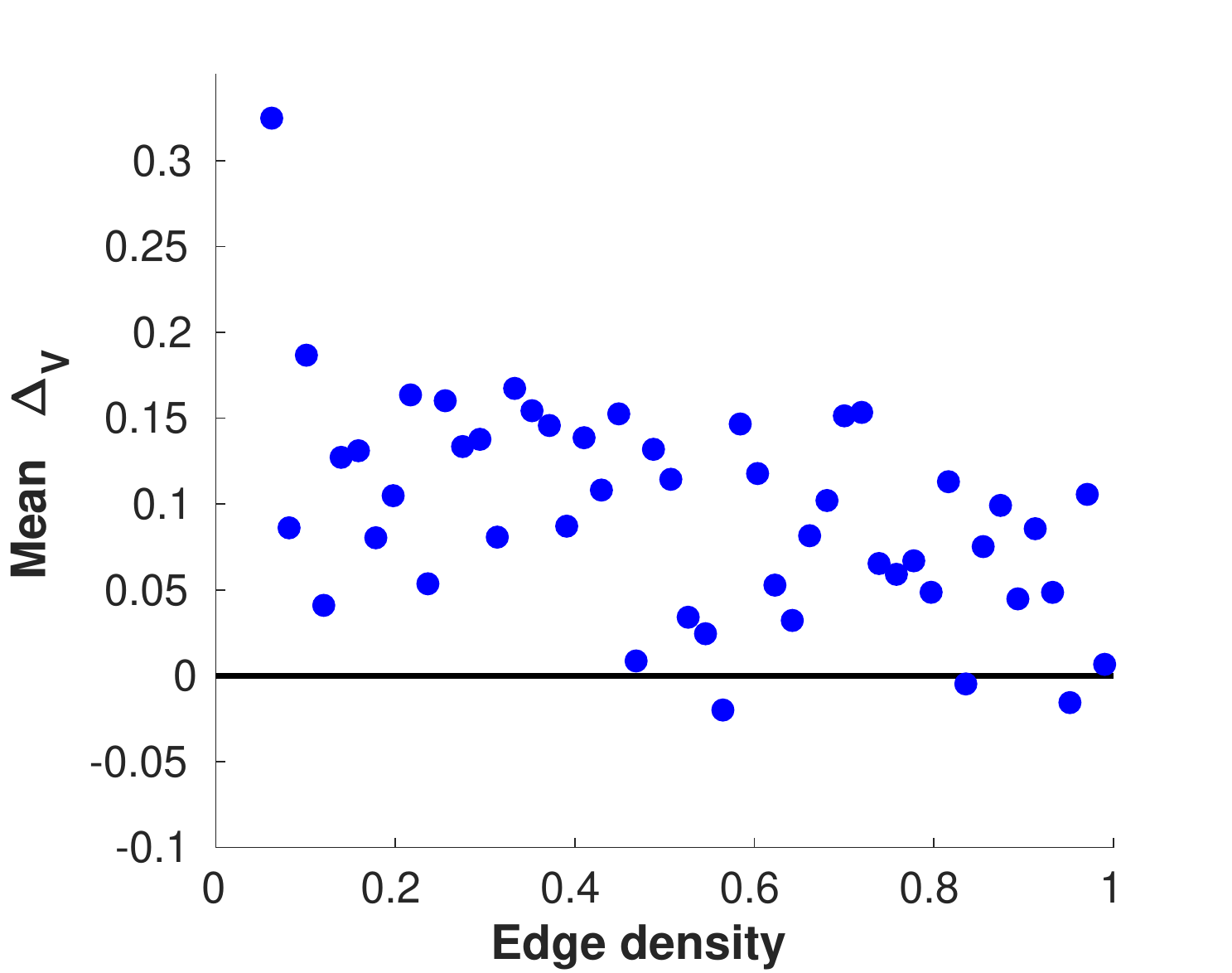}}
			&
			\subcaptionbox{$p_r=50\%$  \label{subfig:valid 50}}{\includegraphics[width=.3\linewidth, height=.16\linewidth]{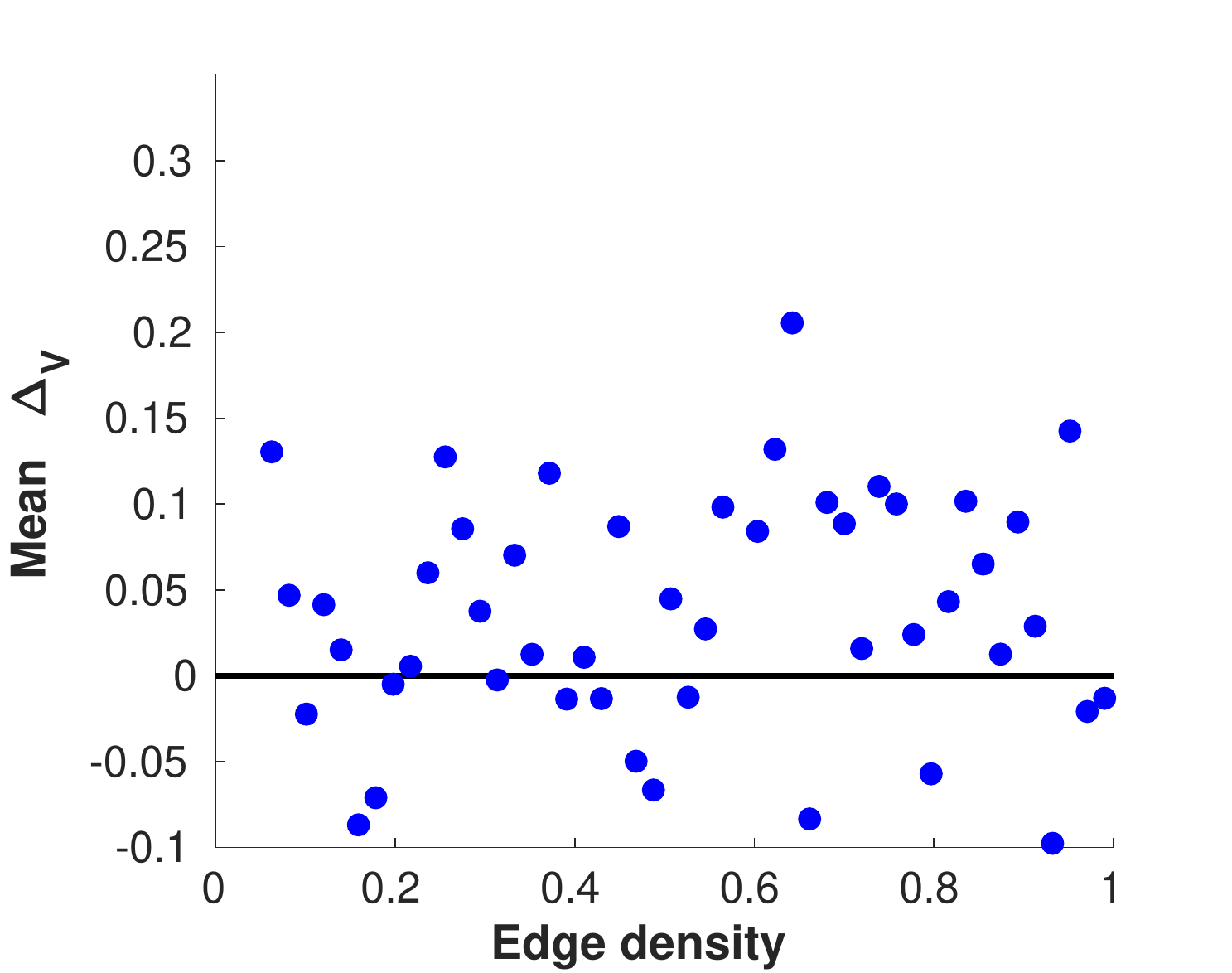}} 
			&		
			\subcaptionbox{$p_r=80\%$  \label{subfig:valid 80}}{\includegraphics[width=.3\linewidth, height=.16\linewidth]{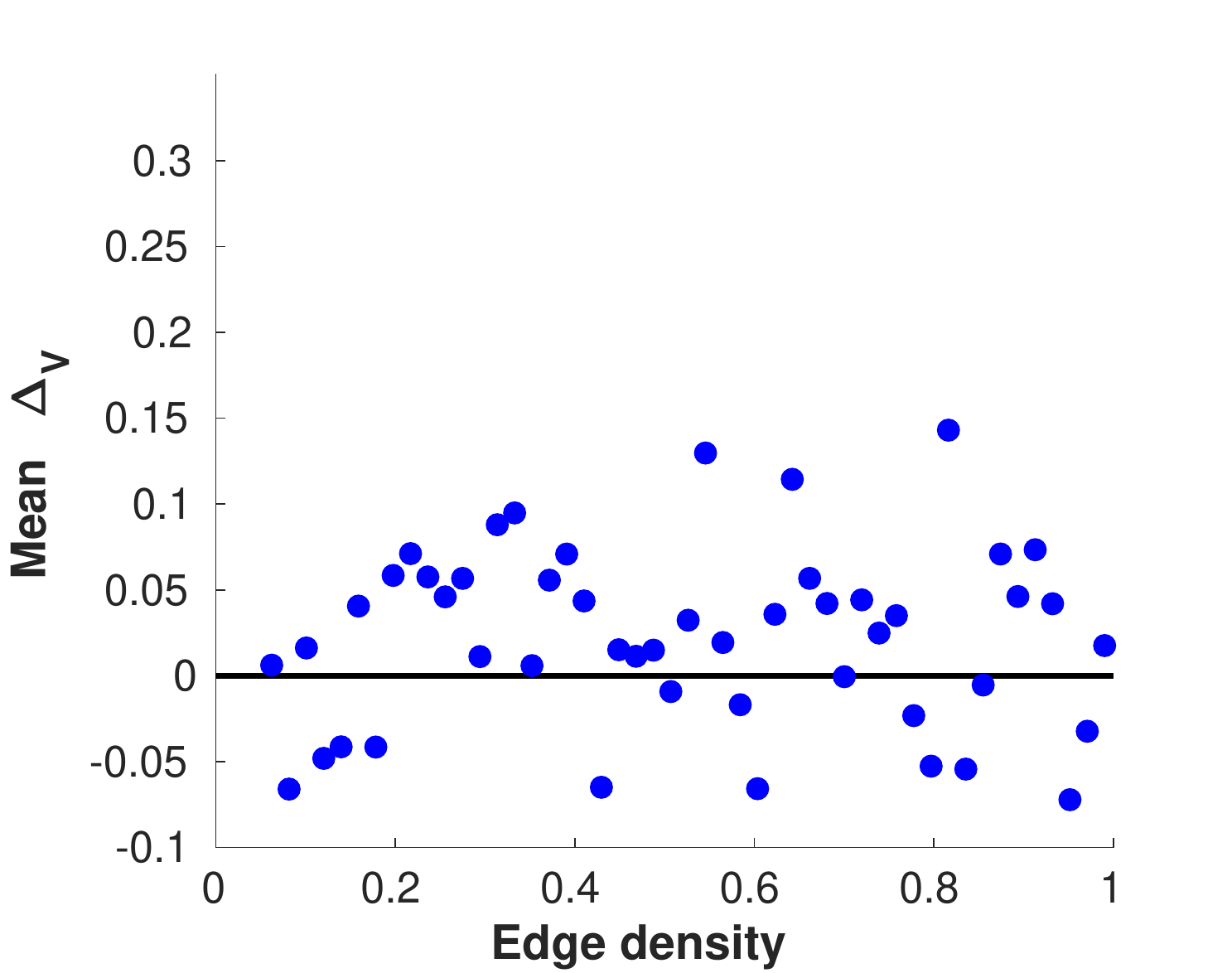}}
		\end{tabular}
		\caption{Comparison of our method against a random edge selection strategy in reducing the task inefficacy of the heterogeneous multi-robot team (\autoref{sec:simulation}).   
			Our method performs significantly better in low resource availability ($p_r=20\%$) scenarios.}  
		\label{fig:sim validation}
	\end{figure*}
	
	
	\begin{figure*}[th]
		\centering
		\begin{tabular}{ccc}
			\centering
			\subcaptionbox{$n=5$   \label{subfig:hind 5}}{\includegraphics[width=.3\linewidth, height=.16\linewidth]{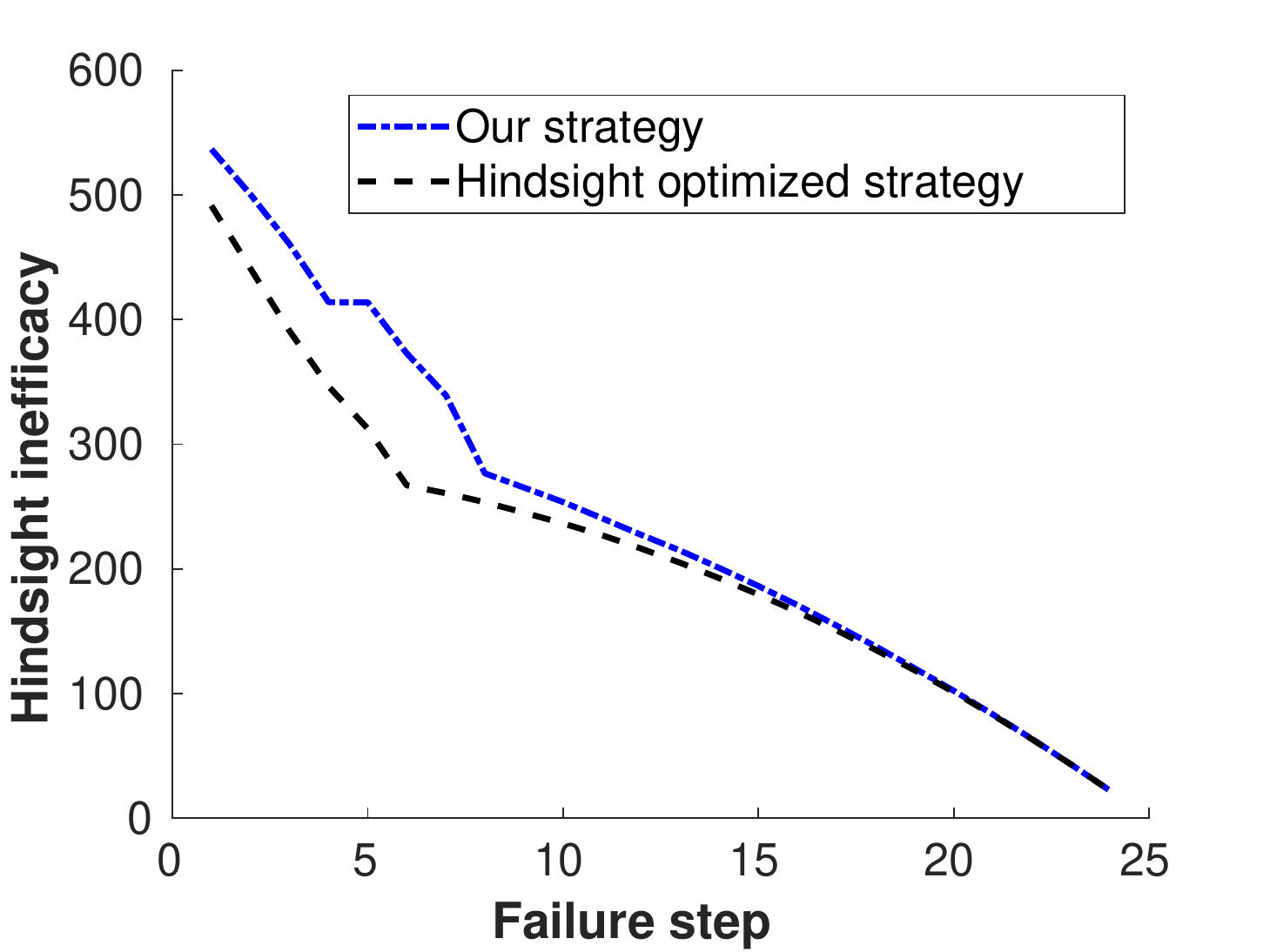}}
			&
			\subcaptionbox{$n=10$  \label{subfig:hind 10}}{\includegraphics[width=.3\linewidth, height=.16\linewidth]{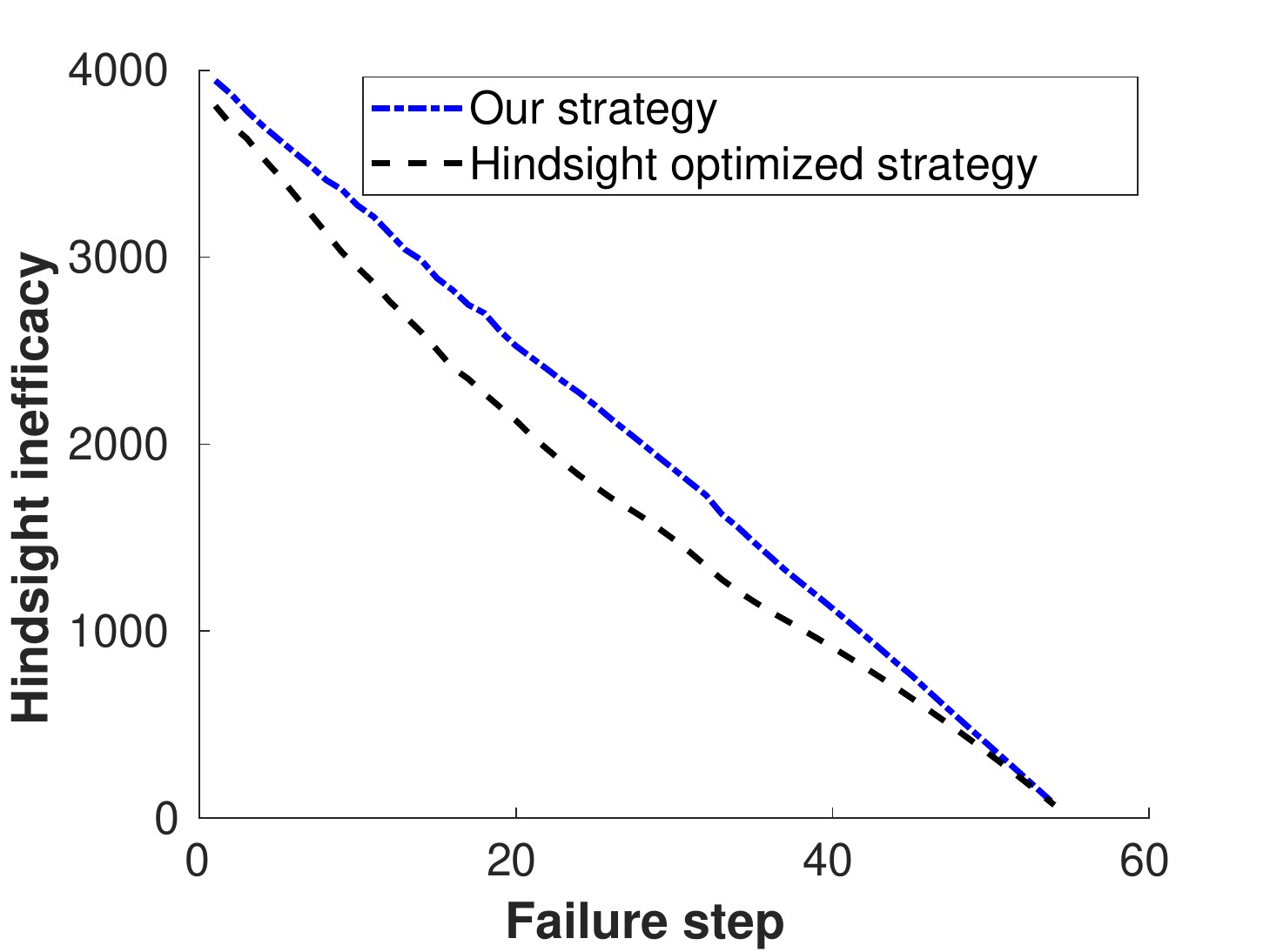}} 
			&		
			\subcaptionbox{$n=20$  \label{subfig:hind 20}}{\includegraphics[width=.3\linewidth, height=.16\linewidth]{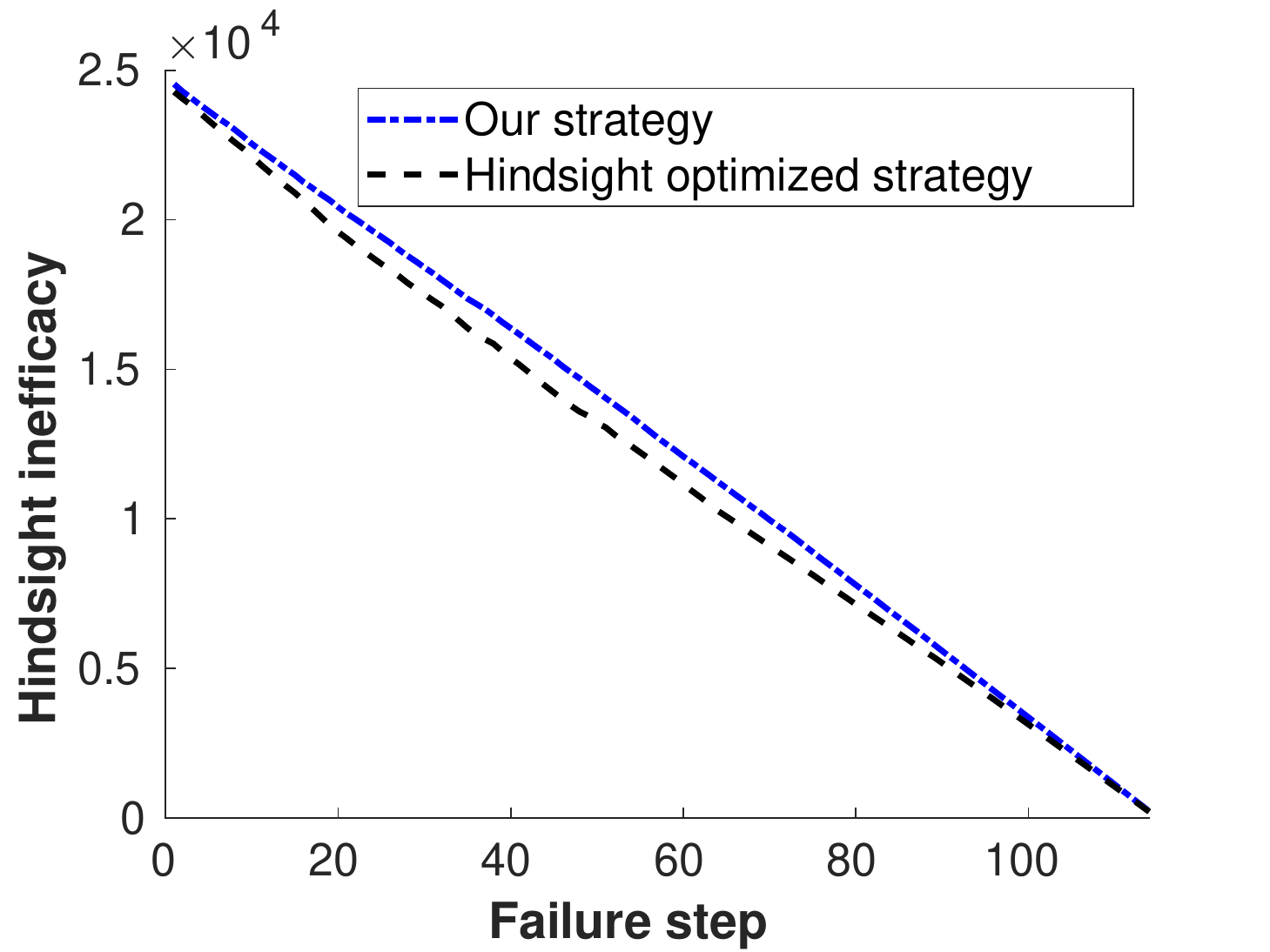}}
		\end{tabular}
		\caption{Comparison of our method against a hindsight optimized algorithm that has access to an oracle.  Blue dotted lines indicate the maximum hindsight inefficacy of the configurations computed using our strategy (over $30$ trials) plotted against the failure steps. Black dashed lines indicate the minimum hindsight inefficacy of the configurations computed using the hindsight optimized strategy (over the $30$ trials) plotted vs. the failure steps for each $n$. }  
		\label{fig:sim hindsight}
		\vspace{-5mm}
	\end{figure*}
	
	We conducted various simulation experiments to validate our approach. For each simulation experiment, the MISDP problem ( \autoref{subsec:config gen}) was solved using MATLAB with YALMIP~\cite{Lofberg04} as the optimization problem modeling interface and SeDuMi~\cite{Sturm_1999} as the semi-definite programming solver. The simulated annealing technique for formation synthesis (\autoref{subsec:Formation synthesis}) was implemented in MATLAB. All simulation and experimental computations were performed on a 64-bit Ubuntu 16.04 desktop with 3.30GHz Intel Core i7-5820K CPU and 32 GB RAM. Parameters used for the simulations were closely matched with corresponding ones used for the tests with physical robots described in \autoref{sec:experiment}. In addition, we choose $c_{min} = d_{s}$, $c_{max} = d_{mc}$ and $ne = 2$. 
	
	As an illustration, \autoref{fig:sim screenshots} depicts screenshots from one simulation trial showing various resource failure instances and the corresponding reconfigurations. This trial was conducted using seven simulated quadrotors, each having three resources at the start of the experiment. As shown in \autoref{subfig:scr_t_0}, the quadrotors' initial positions are predefined such that their communication graph has a line graph topology. A total of $19$ resource failures were simulated before the resource matrix of the team became infeasible. A 3-bit RGB color code is used to represent resources present in each quadrotor. For example, a white color indicates all three resources (RGB for white is $(1,1,1)$). Similarly, a quadrotor rendered in yellow (RGB $(1,1,0)$) represents the case where the quadrotor possesses all the resources except for the third. A quadrotor rendered in black denotes that it has lost all its resources pertinent to the team task. The multimedia attachment submitted with this manuscript shows the video of this simulation. 
	
	In addition, we compared our approach against a strategy in which the robot that suffered from a resource failure connects itself to another random non-communicating robot. We term this the \textit{random edge strategy}.
	For comparison, we randomly generated connected graphs with $n$ between $3$ and $30$. A feasible resource matrix with $r$ between $3$ and $20$ was constructed based on a predefined \textit{resource percentage} $p_r$ for each randomly generated graph. For a given $p_r$, we construct a feasible resource matrix by randomly selecting $\left \lceil \frac{p_r*n*r}{100} \right \rceil$ unique locations from $\mathbf{0}^{n \times r}$ and setting them to $1$, such that, the resultant resource matrix is feasible. We define the \textit{edge density} of a graph with $n$ vertices as the ratio of the number of edges in the graph to the number of edges in a full connected graph with $n$ vertices. We denote $\Delta_{\mathbf{V}}$ as the difference in task efficacy of the new configuration resulting from the random edge addition strategy and our configuration generation strategy. We generated $1000$ random connected graphs and $1000$ feasible resource matrices, for each $p_r \in \{20, 50, 80\}$. \autoref{fig:sim validation} depicts the results from these simulations.
	
	To generate the plots in \autoref{fig:sim validation}, we divide the interval (0,1] into $50$ bins of equal size. The mean $\Delta_{\mathbf{V}}$ of graphs belonging to each bin were computed. The mean $\Delta_{\mathbf{V}}$ of each bin versus the midpoint of the bin interval were plotted. $\Delta_{\mathbf{V}}>0$ implies that our approach has increased the task efficacy more than the random edge strategy. Our strategy is designed to improve task efficacy of the team, so it is bound to improve this quantity for any configuration of interest.
	However, from \autoref{eqn:inefficacy matrix} it is also clear that adding a random edge can not \emph{hurt} task efficacy.
	Since our MISDP uses constraints to enforce that the task efficacy must be improved but does not explicitly maximize the improvement, it is worthwhile to examine how much is gained in comparison with the random edge strategy.
	From the scatter plot \autoref{subfig:valid 20}, since most points lie above zero, we conclude that our strategy increases the task efficacy of the multi-robot team considerably in comparison to the random edge strategy, even when the resource percentage is low. Again in \autoref{subfig:valid 50}, we observe a similar trend in the plot with $p_r=50\%$. Finally, in \autoref{subfig:valid 80}, we do not see a significant difference between the number of points that lie above and below zero. This is expected behavior, since when resource availability in the system is high, the effect of topology of the communication graph on the task efficacy of the multi-robot team is reduced.
	
	We also performed a set of simulations comparing our strategy with a ``hindsight optimized strategy'' which uses information from an oracle about all future failures to compute the new configuration.  To formalize this idea, we define a new metric \textit{hindsight inefficacy}($\|\cdot\|^{\sim}$): $\|\mathbf{\bar{A}}[k]\|^{\sim} = \|n \mathbf{1}^{n\times r} - \mathbf{\bar{A}}[k]\Gamma_r[k]\|_* + \sum_{\tau = k+1}^{h} \|n \mathbf{1}^{n\times r} - \mathbf{\bar{A}}[k]\Gamma_r[\tau]\|_*$ . The hindsight inefficacy matrix quantifies task inefficacy of a given configuration, if  feasible resource matrices associated with a {\em future} tolerable resource failure sequence are obtained through an oracle($[\Gamma_r[k], \Gamma_r[k+1], \cdots, \Gamma_r[h]]$).  For comparison of both strategies, we performed $30$ tolerable failure sequence simulation trials each with $n \in \{5, 10, 20\}$. Each trial was initialized with a line graph and $\mathbf{1}^{n\times 6}$ as the initial communication graph and initial resource matrix respectively. We computed a new configuration for each failure in a tolerable failure sequence using  both our strategy and the hindsight optimized strategy. The hindsight optimized strategy is same as our MISDP problem without the final constraint (\autoref{eqn:cons:reduce inefficacy}), except that it minimizes the hindsight inefficacy instead of the trace of $\mathbf{L}$. Moreover, to quantify the worst case performance, we plotted (\autoref{fig:sim hindsight}) the {\em minimum} hindsight inefficacy of the hindsight optimized configuration and the {\em maximum} hindsight inefficacy of the configuration computed from our strategy over all $30$ trials. \autoref{fig:sim hindsight}, illustrates that the performance of our method matches closely with the hindsight optimized strategy as $n$ increases.
	
	\section{Experimental Results}
	\label{sec:experiment}
	\begin{figure}[ht]
		\centering
		\includegraphics[width=0.8\columnwidth,trim={20mm 20mm 0 78mm},clip]{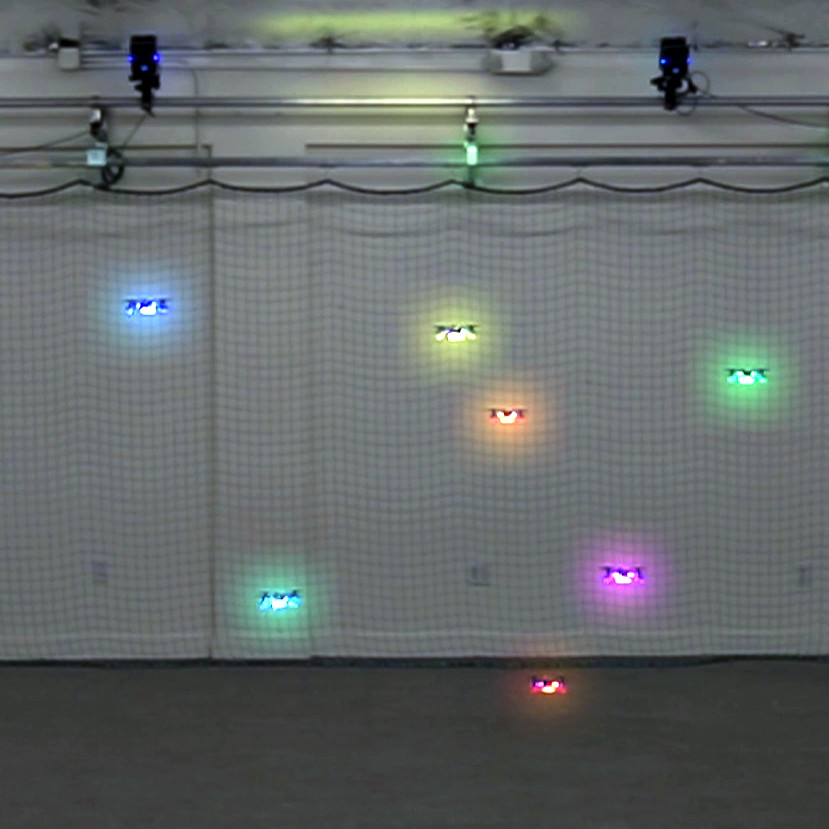}
		\caption{Experiment with seven quadrotors. Each of three resources is mapped to one RGB color channel. A video of the flight trial is included in the supplementary material.}
		\vspace{-3mm}
	\end{figure}
	We additionally demonstrate our approach using the Crazyswarm multi-quadrotor experimental platform~\cite{crazyswarm} with a team of seven small quadrotors.
	We execute the same sequence shown in the simulator screenshots in \autoref{fig:sim screenshots}.
	This illustrates that the formation synthesis and trajectory planning stages of our method produce results that satisfy the collision and kinodynamic constraints of a multi-quadrotor team in the real world.
	In our physical setup, due to programming language interoperability issues we precomputed the sequence of configurations, formations, and multi-robot trajectory plans. However the running time of our method is fast enough for real-time use:
	in our example problem with seven robots, the average computation time was \SI{0.66}{\second} for configuration generation, \SI{2.43}{\second} for formation generation, and \SI{0.15}{\second} for motion planning, resulting in a total of \SI{3.24}{\second}.
	We note that our implementation of simulated annealing neglects opportunities for performance optimization in evaluating $E(x)$ after updating a single coordinate, and we believe a well-tuned implementation could be significantly faster.

	\section{Conclusions}
	\label{sec:conclusion}
	We described a novel method that enables a heterogeneous team of robots performing a task to reconfigure themselves to a new formation in the event of a robot resource failure. The new formation reduces the impact of resource failure in the team by shifting the robots to a new configuration such that the resultant configuration is closer to an ideal configuration without adding significant communication cost (number of edges) to the network.  Our method has two parts. The first (posed as a MISDP) generates the new configuration in the form of a communication graph and the inter-robot distance between the communicating robots. The second part, (optimized with simulated annealing) computes the robots' spatial coordinates to realize the configuration generated in the previous part in 3D. We validated our strategy through simulations and demonstrated it on multiple quadrotors.  Future work involves incorporating explicit models of resources, namely sensing models, actuation models, and internal computation models, into the strategy and decentralizing the strategy. 
	

	\addtolength{\textheight}{-1cm}   
	


	%
	
	
	
	

\end{document}